\pdfoutput=1

\documentclass[11pt]{article}

\usepackage[preprint]{acl}

\usepackage{times}
\usepackage{latexsym}

\usepackage[T1]{fontenc}

\usepackage[utf8]{inputenc}

\usepackage{microtype}

\usepackage{inconsolata}

\usepackage{graphicx}

\usepackage{hyperref}
\usepackage{url}
\usepackage{soul}
\usepackage{pifont}
\usepackage{xcolor}
\usepackage{colortbl}
\usepackage[most]{tcolorbox}
\usepackage{algorithm}
\usepackage{pifont}
\usepackage[noend]{algpseudocode}
\usepackage{amsfonts}
\usepackage{graphicx}
\usepackage{subfig}
\usepackage{mathtools}
\usepackage{amsmath}
\usepackage{amsthm}
\usepackage{listings}
\usepackage{wrapfig}
\usepackage{makecell}
\definecolor{light_gray}{rgb}{.95,.95,.95}
\definecolor{custompurple}{RGB}{93,0,93}
\definecolor{customorange}{RGB}{255,132,6}
\definecolor{customgold}{RGB}{213,177,52}
\definecolor{customblue2}{RGB}{28,205,188}
\definecolor{no_persona_color}{RGB}{152,226,245}
\definecolor{persona_color}{RGB}{193,167,246}
\definecolor{customorange}{RGB}{230, 130, 30}
\definecolor{customblues3}{RGB}{70, 130, 180}
\definecolor{wiserblue}{HTML}{DCEEFD}
\definecolor{wiseryellow}{HTML}{FFF3CC}
\usepackage{booktabs}
\usepackage{adjustbox}
\usepackage{scalerel}
\usepackage{fontawesome5}
\newtcolorbox{policyPrompt}{
  enhanced,
  breakable,
  colback=promptbg,
  colframe=promptframe,
  boxrule=0.4pt,
  arc=1pt,
  left=6pt,
  right=6pt,
  top=6pt,
  bottom=6pt,
  fontupper=\ttfamily\small,
  coltitle=promptaccent,
  title={\textbf{Action Selection Prompt Structure}}
}

\definecolor{myblue}{RGB}{30,92,180}
\definecolor{myRed}{RGB}{180,40,40}

\newtcolorbox{promptblock}{colback=gray!5!white,colframe=gray!75!black,
  title=Prompt Block, breakable, fontupper=\ttfamily\small}

\title{PerceptUI: LLM Agents as Human-Aligned Synthetic Users for UI/UX Evaluation}

\author{
 \textbf{Nicolas Bougie\textsuperscript{1}},
 \textbf{Xiaotong Ye\textsuperscript{1}},
 \textbf{Gian Maria Marconi\textsuperscript{1}},
 \textbf{Narimasa Watanabe\textsuperscript{1}}
 \\ \texttt{\{nicolas.bougie,tony.yip,gianmaria.marconi,narimasa.watanabe\}@woven.toyota}\\
\\
 \textsuperscript{1}Woven by Toyota
}

\begin{document}
\maketitle
\begin{abstract}
User interface (UI) and user experience (UX) evaluation is central to product development, yet reliable feedback still relies on recruiting human participants or running online A/B tests, making early-stage iteration slow and costly. In light of this, recent work has explored Multimodal Large Language Models as proxy evaluators. However, existing approaches either produce surface-level critiques or a judgment that reflects the model's own biases rather than the genuine response of a particular user. We introduce \textsc{PerceptUI}, a framework for persona-conditioned UI/UX evaluation that predicts how a specific user would answer interface-related questions and produces natural-language rationales. \textsc{PerceptUI} is trained in two stages: (i) \emph{contrastive reflection} fine-tuning distills teacher-generated rationales by extracting lessons from human decisions, and (ii) a reflective prompt-evolution step from the model's own failure traces. Across multiple domains and datasets, \textsc{PerceptUI} achieves human-level realism, generalizes to unseen questions and personas, and yields population-level response distributions.
\end{abstract}

\section{Introduction}
User interface design shapes how users perceive, navigate, and engage with digital interfaces. Small changes in layout, wording, visual hierarchy, or icons can lead to measurable differences in experience, preference, and behavior \citep{fogg2002persuasive,wu2024uiclip,jeon2025wiserui}. Evaluating these effects usually requires user studies or A/B experiments, where participants answer questions about a product or choose between design variants. Collecting such evidence is valuable but costly, resulting in many interfaces being released with limited formal evaluation \citep{luera2025mlljudge,tan2026avenir}.

Recent work has explored using Multimodal Large Language Models (MLLMs) as automatic evaluators. One line of work predicts interface quality from screenshots, either by training models on synthetically degraded UI pairs \citep{wu2024uiclip} or by prompting MLLMs with design heuristics \citep{duan2024autofeedback,duan2024uicrit,hwang2024vpir}. Another line studies whether vanilla MLLMs can predict average human preferences between two UI variants \citep{jeon2025wiserui,luera2025mlljudge}. While these methods provide useful signals, they usually evaluate interfaces from a model-centric perspective. Few-shot prompting can reflect the MLLM's own preferences or biases, while supervised fine-tuning on answers does not explicitly model \emph{why} a particular user would select one answer over another, limiting agent's ability to generalize to new interfaces, questions, and user profiles. This drawback is critical because UI judgments are often user-dependent. The same interface may be perceived differently depending on a user's goals, prior experience, visual preferences, or domain familiarity. A useful surrogate evaluator should therefore go beyond predicting an average preference and simulate how a specific user would respond to a given UI/UX evaluation question.

\begin{figure}[tbp]
    \centering
    \includegraphics[width=1.0\linewidth]{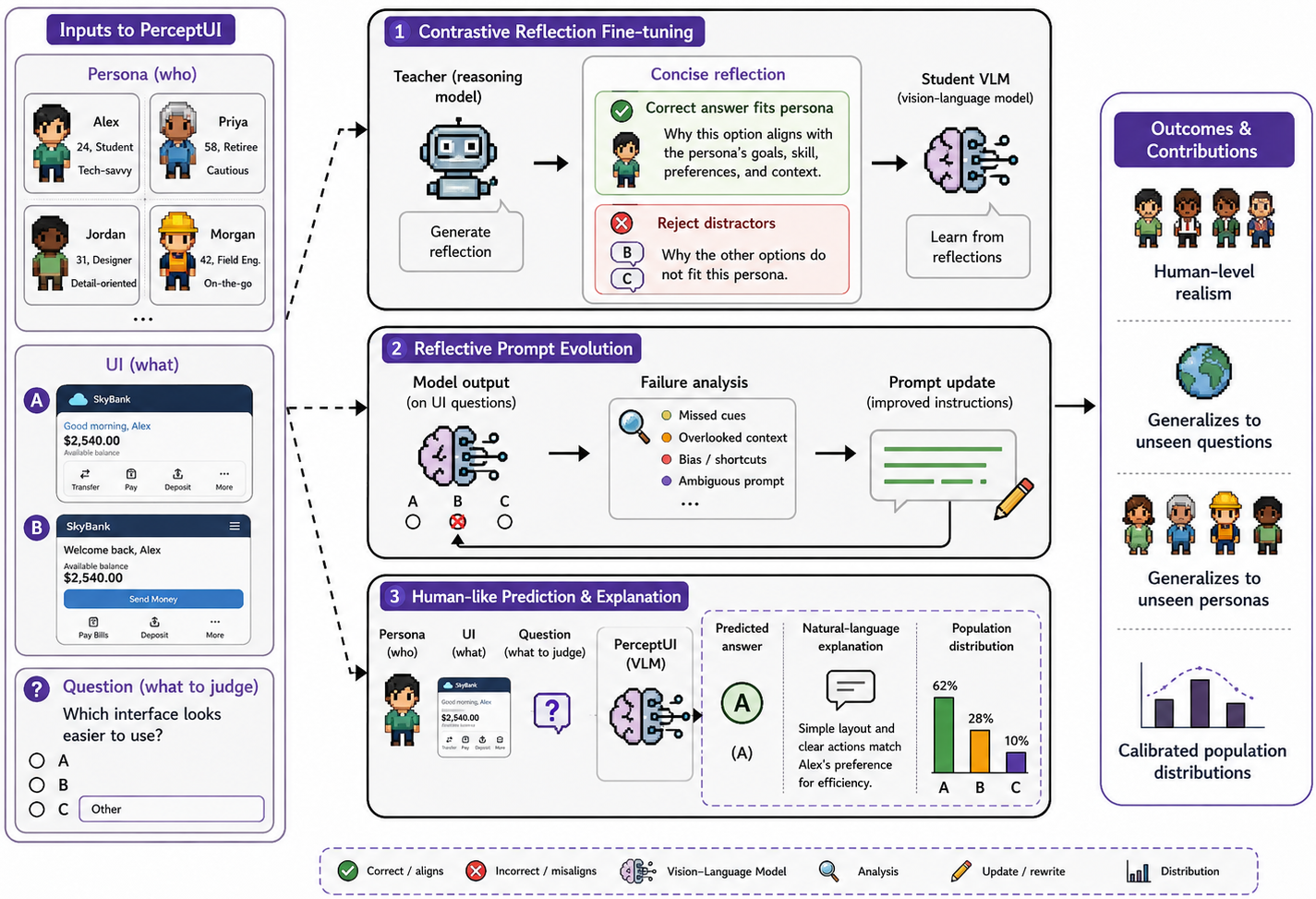}
    \caption{Overview of \textsc{PerceptUI}, a framework for persona-conditioned UI/UX evaluation.}
    \label{fig:overall_method}
\end{figure}

We introduce \textsc{PerceptUI}, a framework for persona-conditioned UI/UX evaluation. Given a UI screenshot, a user persona, and a question, \textsc{PerceptUI} predicts how the corresponding user would answer and generates a natural-language rationale for the prediction. Our approach is motivated by two observations. First, imitation learning is a weak training signal as it tells the model \emph{what} to output but not \emph{why} the correct option fits, and the other options do not. We therefore leverage a teacher model to turn each example into a \emph{contrastive reflection} that justifies both the chosen answer and the rejection of each alternative, and use these traces to fine-tune a vision-language model. Second, evaluation questions vary substantially across surveys in wording, answer scales, and required interpretation. Thus, we apply a reflective prompt-evolution step, where errors made by the model are summarized and used to refine the evaluation prompt. This step adapts the inference prompt to recurring failure patterns without relying on manual survey-specific prompt tuning. Across multiple UI/UX evaluation tasks, \textsc{PerceptUI} reaches human-level performance, generalizes to unseen questions and personas, and produces calibrated population-level answer distributions. Overall, the framework supports both early-stage evaluation and targeted analysis of how different designs affect different users.

\section{Related Work}

\textbf{UI Evaluation with MLLMs.} Automatic UI evaluation has long studied computational measures of visual complexity, aesthetic structure, and layout preference from interface properties \citep{miniukovich2015complexity,reinecke2013pref}. More recent work uses vision-language models to evaluate UI screenshots directly. One line of work trains dedicated models for design-quality prediction. For example, UIClip~\citep{wu2024uiclip} fine-tunes CLIP on BetterApp, a dataset of designer-rated mobile UI comparisons. Another line of work prompts MLLMs to produce natural-language critiques from design heuristics or visual evidence. \citep{duan2024autofeedback} query GPT-4 with Nielsen-style heuristics, UICrit~\citep{duan2024uicrit} studies critique generation from expert mobile-UI feedback, and \citep{hwang2024vpir} improve grounding through region-level visual prompts. Beyond single-model critique, recent systems explore broader forms of MLLM-assisted design evaluation, including heuristic-evaluation reproduction \citep{mendes2025gpt4oheur,rossi2025synthetic}, concrete UI-fix recommendation \citep{kim2025recommending}, multi-role critique \citep{chen2026critiquecrew}, UX-metric selection \citep{duan2024evalign}, and feedback-based improvement of UI generation models \citep{duan2026improvingui}. However, these approaches mainly produce generic judgments or critiques, without modeling why a specific human response follows from the UI, question, and user context.

\noindent\textbf{MLLM-as-Judge Benchmarks.} A complementary line of work evaluates MLLMs as judges for UI/UX assessment. WiserUI-Bench~\citep{jeon2025wiserui} tests pairwise UI selection using A/B-tested variants with validated winners and expert interpretations, demonstrating that current MLLMs remain sensitive to position bias and unreliable at predicting behavior-aligned preferences. Similarly, MLLMJudge \citep{luera2025mlljudge} compares MLLM judgments with human ratings of web interfaces. These benchmarks are valuable because they use human judgments or behavioral outcomes as the target signal, rather than relying only on expert heuristics or synthetic design defects. Our work also leverages human answers as supervision. \textsc{PerceptUI} builds on this human-centered principle, while explicitly supervising the model to predict how individual users respond and why.

\noindent\textbf{LLM Agents as Synthetic Users.} LLM agents have increasingly been used to approximate human behavior across domains \citep{gao2023s, bougie2026beyond, bougie2026alignuser, bougie2025citysim, zhang2023generative}.  Close to UI evaluation, SimUser \citep{xiang2024simuser} simulates mobile-app users to surface heuristic usability issues, UXAgent \citep{lu2025uxagent,lu2025uxagentshort} runs persona-conditioned agents on live web targets, and AgentA/B \citep{lu2025agentab} deploys $1{,}000$ agents on Amazon to approximate A/B testing outcomes. UXCascade \citep{holter2026uxcascade} aggregates agent traces into usability findings, Avenir-UX \citep{tan2026avenir} grounds interaction in GUI observations and produces SUS/SEQ-style reports, and SimAB \citep{holter2026simab} predicts outcomes for historical A/B experiments. Related work on LLM agents studies accessibility, reliability, and web interaction behavior \citep{taeb2024axnav,peng2026uxagents,wang2025webprober}. OPeRA \citep{wang2025opera} records online-shopping sessions that pair user personas with observed interfaces, actions, and rationales, and \citep{kim2025synthcogwalk} aligns LLMs with human cognitive walkthroughs. 

\noindent\textbf{Reasoning Distillation and Prompt Optimization.}
Our method also relates to reasoning distillation and prompt optimization. Prior work distills chain-of-thought rationales from stronger teachers into smaller students to improve reasoning \citep{wei2022cot,hsieh2023distilling}. We adapt this idea to subjective UI/UX evaluation by supervising the model with contrastive rationales: each rationale explains why the recorded human answer fits the UI and persona better than the distractors. This follows the motivation of contrastive explanations, which clarify why one outcome occurred instead of another \citep{miller2019explanation}. At inference time, we build on reflective prompt-optimization methods such as GEPA and TextGrad, which revise prompts from failure traces \citep{agrawal2025gepa}. Our reflective prompt evolution applies this principle to UI/UX evaluation by updating the UI-grounding, persona-grounding using observed failures.

\section{Method}
\label{sec:method}
\subsection{Problem Formulation}
We formulate UI/UX evaluation as persona-conditioned question answering. Each example is represented as a tuple $x_i = (s_i, r_i, q_i, \mathcal{Y}_i, p_i)$, where $s_i$ is a UI screenshot, $r_i$ is an optional reference image such as an alternative icon or UI variant, $q_i$ is a question, $\mathcal{Y}_i=\{y_{i1},\ldots,y_{iK}\}$ is the set of answer options, and $p_i$ is a natural-language persona describing participant $i$. Given $x_i$ and an instruction prompt $P$, the model generates both a rationale $z_i$ and an answer $y_i$. The joint distribution can be formalized as: $p_\theta(z_i, y_i \mid s_i, r_i, q_i, \mathcal{Y}_i, p_i; P)$, where $\theta$ are the trainable parameters and $P$ is task-specific prompt, with the answer marginal expressed as: $p_\theta(y \mid \cdot) = \sum_{z} p_\theta(z, y \mid \cdot)$. We assume that the ground truth is the answer $y^\star \in \mathcal{Y}$. Averaging the answer marginal over a population of personas $\{p^{(n)}\}_{n=1}^{N}$, the model induces an answer distribution: $\hat{p}(y \mid s, r, q; P) = \frac{1}{N}\sum_{n=1}^{N} p_\theta(y \mid s, r, q, \mathcal{Y}, p^{(n)}; P)$, which we can compare against the empirical distribution of human answers to evaluate a candidate design. In this work, $p_\theta$ denotes the MLLM used for persona-conditioned answer prediction. To align $p_\theta$ with genuine users, we adopt a two-stage pipeline: contrastive reflection fine-tuning teaches the model to disentangle human preferences, and reflective prompt evolution addresses residual errors.

\subsection{Contrastive Reflection Fine-Tuning}
A standard supervised objective trains the model to imitate the human answer $y_i^\star$, but provides no direct supervision about \texttt{why} that answer is appropriate or why the alternatives are less suitable. This is limiting for subjective UI evaluation, where the correct response often depends on the interaction between visual evidence, question wording, and persona-specific preferences or constraints. Thus, we introduce contrastive reflection, a mechanism that enables agents to identify the reasons behind decisions and relate them to their internal context.

\noindent For each training example $x_i=(s_i,r_i,q_i,\mathcal{Y}_i,p_i)$, we first construct task-conditioned auxiliary context. A UI grounding prompt extracts question-relevant visual evidence from the screenshot and optional reference image, yielding a summary $u_i$. A persona grounding prompt maps the participant description to preferences, constraints, or sensitivities relevant to the question and answer options, producing a persona summary $g_i$. Both summaries are generated without access to the ground-truth answer $y_i^\star$ and are appended to the answer prompt. Conditioned on this augmented input and the recorded answer, a teacher model $T$ generates a contrastive rationale:
\begin{equation}
c_i^\star =
T(s_i,r_i,q_i,\mathcal{Y}_i,p_i,u_i,g_i,y_i^\star).
\end{equation}
We structure $c_i^\star$ into three short components: \emph{UI evidence}, \emph{persona relevance}, and \emph{option contrast}. The persona-relevance component explains which aspects of the participant profile matter for the decision, if any. The option-contrast component justifies the recorded answer $y_i^\star$ while contrasting it with each distractor $y_{ik}\in\mathcal{Y}_i\setminus\{y_i^\star\}$.

\noindent Then, we fine-tune the student model $p_\theta$ to generate the contrastive rationale followed by the recorded answer:
\begin{equation}
\resizebox{1.0\columnwidth}{!}{$
\mathcal{L}_{\mathrm{CRFT}}(\theta)
= - \mathbb{E}_{(x_i,y_i^\star,c_i^\star)}
\left[
\log p_\theta
\left(
c_i^\star, y_i^\star
\mid
s_i,r_i,q_i,\mathcal{Y}_i,p_i,u_i,g_i;P_{\mathrm{train}}
\right)
\right]
$}
\label{eq:crft_loss}
\end{equation}
Unlike standard rationale distillation \citep{hsieh2023distilling}, which typically supervises a single positive reasoning trace, our target is answer-contrastive. The model is trained to justify the human answer relative to the full option set, encouraging UI-grounded and persona-aware comparisons rather than label imitation alone.

\subsection{Reflective Prompt Evolution}
\label{sec:rpe}
After contrastive reflection fine-tuning, we employ the student model $p_\theta$ as a synthetic user. Although contrastive reflection fine-tuning improves the model's ability to compare UI evidence and persona-specific preferences, a generic prompt may generalize poorly across surveys and some residual errors may remain. We therefore optimize the prompt $P$ using reflective prompt evolution. During this stage, the model parameters $\theta$ remain fixed and only the prompt is optimized.

Starting from an initial prompt $P_0$, each iteration evaluates the current prompt $P_t$ on a minibatch $\mathcal{B}_t \subset \mathcal{D}_{\mathrm{dev}}$. For each example, we collect the generated UI summary $u_i$, persona summary $g_i$, rationale $c^{*}_i$, predicted answer $\hat{y}_i$, and the corresponding human target. For population-level tasks, we aggregate predictions across personas to obtain the predicted answer distribution $\hat{p}_t(\cdot \mid s_i,r_i,q_i,\mathcal{Y}_i)$. A prompted evaluator $E$ compares this distribution with the empirical human distribution and produces a symbolic loss:
\begin{equation}
\mathcal{L}_{\mathrm{sym}}^{(t)}
=
E(P_t,\mathcal{B}_t).
\label{eq:rpe_loss}
\end{equation}
The symbolic loss records the signed residuals between predicted and empirical answer probabilities, together with representative failure traces. The screenshot and reference image are included in the evaluator input so that the critique remains grounded in the UI.

A second LLM analyzer $A$ analyzes these failures and identifies weaknesses in the current prompt, such as over-reliance on generic persona traits, insufficient attention to the answer options, or weak use of the intermediate summaries. This produces a symbolic gradient:
\begin{equation}
\nabla_{\mathrm{sym}}^{(t)}
=
A(P_t,\mathcal{L}_{\mathrm{sym}}^{(t)}),
\label{eq:rpe_grad}
\end{equation}
which specifies prompt-level edits likely to reduce the observed residuals. A prompt optimizer $O$ then applies these edits to produce a revised prompt:
\begin{equation}
P'_t = O(P_t,\nabla_{\mathrm{sym}}^{(t)}).
\label{eq:rpe_update}
\end{equation}
Each candidate prompt is evaluated on $\mathcal{D}_{\mathrm{dev}}$ and added to a candidate pool. Before final selection, an LLM judge audits each candidate prompt and rejects prompts that contain answer leakage, dataset-specific shortcuts, or hard-coded decision rules. After $T$ rounds, we return $P^\star$, the candidate with the lowest mean residual on $\mathcal{D}_{\text{dev}}$.

\subsection{Backbone and Inference}
\label{sec:backbone_inference}

We instantiate $p_\theta$ with Qwen-VL. The model parameters $\theta$ are updated only during contrastive reflection fine-tuning and are kept fixed during reflective prompt evolution. At inference, we employ the evolved prompt $P$. Given an input $x=(s,r,q,\mathcal{Y},p)$, the model first generates a UI grounding summary $u$ and a persona grounding summary $g$, and then predicts a rationale and answer:
\begin{equation}
(\hat{z},\hat{y})
\sim
p_\theta(
\cdot
\mid
s,r,q,\mathcal{Y},p,u,g;P^\star
).
\end{equation}

\section{Experiments}

\textbf{Settings.}
We evaluate \textsc{PerceptUI} on six datasets covering complementary UI/UX tasks: WiserUI-Bench~\citep{jeon2025wiserui}, UIClip/BetterApp~\citep{wu2024uiclip}, WebDevJudge~\citep{luera2025webdevjudge}, LabintheWild ~\citep{reinecke2014quantifying}, LabintheWild-UX ~\citep{miniukovich2024prototypicality}, and UICrit~\citep{duan2024uicrit}. We additionally evaluate on a proprietary survey that captures how users perceive and interpret in-car interfaces, UXCar. For datasets without user profiles, we omit the persona field from the prompt and evaluate \textsc{PerceptUI} as a non-personalized UI evaluator. 

\noindent\textbf{Baselines.} We compare \textsc{PerceptUI} against the baselines reported in each benchmark. For WiserUI-Bench, these include proprietary and open MLLMs, together with reasoning-enhanced methods such as CoCoT. For the remaining public benchmarks, we follow the original evaluation protocols and report the strongest baselines.

\subsection{UI/UX Design Selection}
\label{sec:exp_wiserui_selection}

\begin{table}[tbp]
\centering
\small
\resizebox{1.0\linewidth}{!}{
\begin{tabular}{llcccc}
\toprule
\textbf{Model} & \textbf{Setting} & \textbf{FA}$\uparrow$ & \textbf{SA}$\uparrow$ & \textbf{AA}$\uparrow$ & \textbf{CA}$\uparrow$ \\
\midrule
Random & -- & 50.00 & 50.00 & 50.00 & 25.00 \\
\midrule
o1 & Zero-shot & 16.56 & \cellcolor{wiserblue}\textbf{97.78} & 57.17 & 15.56 \\
GPT-4o & Zero-shot & 31.89 & 88.33 & 60.11 & 30.11 \\
GPT-5 & Zero-shot & 33.91 & \cellcolor{wiseryellow}\underline{89.80} & 61.86 & 31.09 \\
Claude Opus 4.6 & Zero-shot & 28.13 & 84.55 & 56.34 & 26.44 \\
Qwen2.5-VL-32B & Zero-shot & 33.67 & 81.44 & 57.56 & 31.00 \\
InternVL-2.5-38B & Zero-shot & 55.67 & 60.00 & 57.83 & 34.56 \\
LLaVA-NeXT-7B & Zero-shot & 54.44 & 40.56 & 47.50 & 10.78 \\
LLaVA-OneVision-7B & Zero-shot & 19.78 & 81.56 & 50.67 & 10.44 \\
\midrule
GPT-5 & CoCoT & 35.22 & 86.14 & 60.68 & 32.78 \\
GPT-5 & Self-Refine & 34.00 & 85.34 & 59.67 & 31.44 \\
GPT-5 & DDCoT & 38.13 & 87.87 & 63.00 & 36.20 \\
GPT-5 & MAD (R1) & 53.78 & 69.62 & 61.70 & 40.56 \\
Claude Opus 4.6 & MAD (R1) & 42.33 & 74.07 & 58.20 & 33.11 \\
\midrule
\textsc{PerceptUI} & w/o CR  & 37.40 & 73.80 & 55.60 & 37.20 \\
\textsc{PerceptUI} & w/o RPE  
& \cellcolor{wiseryellow}\underline{59.80} 
& 85.20 
& \cellcolor{wiseryellow}\underline{72.50} 
& \cellcolor{wiseryellow}\underline{41.10} \\
\rowcolor{blue!8}
\textsc{PerceptUI} & - 
& \cellcolor{wiserblue}\textbf{62.10} 
& 86.40 
& \cellcolor{wiserblue}\textbf{74.25} 
& \cellcolor{wiserblue}\textbf{44.30} \\
\bottomrule
\end{tabular}}
\caption{UI/UX selection on WiserUI-Bench. All metrics are accuracy-based; higher is better. FA and SA measure accuracy when the A/B-test winner appears first or second, AA is their average, and CA measures order-consistent correctness. Best and second-best scores are highlighted in blue and yellow.}
\label{tab:wiserui_selection}
\end{table}
A central task for UX research is selecting the best design. We therefore evaluate UI/UX selection on WiserUI-Bench~\citep{jeon2025wiserui}, which contains 300 UI pairs with A/B-test-validated winners. Given two UI variants, the model is instructed to select the design more likely to guide users toward the intended action. We report First Accuracy (FA), Second Accuracy (SA), Average Accuracy (AA), and Consistent Accuracy (CA). CA is the most important metric as it requires the model to select the correct UI under both input orders, reducing the effect of position bias. Table~\ref{tab:wiserui_selection} illustrates that existing MLLMs remain sensitive to input order. Several baselines obtain high SA but much lower FA, indicating a tendency to select the second image rather than consistently identify the A/B-test winner. \textsc{PerceptUI} improves the order-robust metrics, achieving the best AA and CA among all methods. The gain is most pronounced on CA, suggesting that \textsc{PerceptUI} better captures behavior-relevant UI evidence rather than relying on positional cues.

\subsection{UI Design Quality}
\label{sec:exp_uiclip_design}

\begin{table}[tbp]
\centering
\small
\resizebox{0.88\linewidth}{!}{
\begin{tabular}{llc}
\toprule
\textbf{Model} & \textbf{Setting} & \textbf{Overall Acc.}$\uparrow$ \\
\midrule
GPT-4V & Zero-shot & 51.58 \\
GPT-5 & Zero-shot & 55.36 \\
Claude Opus 4.6 & Zero-shot & 53.71 \\
LLaVA-1.6-13B & Zero-shot & 33.20 \\
\midrule
UIClip & CLIP pre-train only & 65.42 \\
UIClip & JitterWeb + Web Pairs + Human & 73.88 \\
UIClip & JitterWeb + Web Pairs & 75.12 \\
\midrule
\textsc{PerceptUI}  & w/o CR & 39.55 \\
\textsc{PerceptUI}  & w/o RPE & \cellcolor{wiseryellow}\underline{76.89} \\
\rowcolor{blue!8}
\textsc{PerceptUI} & - & \cellcolor{wiserblue}\textbf{79.28} \\
\bottomrule
\end{tabular}}
\caption{UI design preference prediction on UIClip/BetterApp. Given a pair of UI screenshots, the task is to select the higher-quality design.}
\label{tab:uiclip_design}
\end{table}
We next evaluate UI quality on the UIClip dataset ~\citep{wu2024uiclip}. Unlike WiserUI-Bench, which tests whether a model can identify the A/B-test winner, UIClip/BetterApp focuses on visual design quality. Given two UI screenshots, the model selects the interface judged to have higher design quality. Table~\ref{tab:uiclip_design} depicts that zero-shot LVLMs perform close to chance, while UIClip variants benefit substantially from UI-specific training data. \textsc{PerceptUI} achieves the highest overall accuracy, with a modest gain over the strongest UIClip variant. This suggests that \textsc{PerceptUI} can act as a general UI evaluator even when persona information is unavailable, while also showing that specialized UI contrastive pre-training remains robust for static design-quality assessment.

\subsection{Human Evaluation of UI/UX Rationales}
\label{sec:exp_rationale_quality}
\begin{table}[tbp]
\centering
\small
\resizebox{0.98\linewidth}{!}{
\begin{tabular}{lcccc}
\toprule
\textbf{Method} & \textbf{UI Grounding} & \textbf{Persona Use} & \textbf{Contrastiveness} & \textbf{Overall} \\
\midrule
GPT-4o & 3.42 & 3.05 & 3.11 & 3.28 \\
GPT-5 & 3.61 & 3.32 & 3.47 & 3.56 \\
Gemini-2.5-pro & 3.26 & 2.88 & 3.02 & 3.17 \\
Claude Opus 4.6 & 3.34 & 3.01 & 3.13 & 3.29 \\
\midrule
SFT & 2.61 & 2.38 & 2.29 & 2.74 \\
\textsc{PerceptUI} w/o CR & 1.70 & 1.55 & 1.61 & 1.89 \\
\textsc{PerceptUI} w/o RPE & 3.68 & 3.56 & 3.61 & 3.71 \\
\rowcolor{blue!8}
\textsc{PerceptUI} & 3.91 & 3.74 & 3.88 & 3.94 \\
\bottomrule
\end{tabular}}
\caption{Human evaluation of rationale quality on UXcar. Annotators rate explanations on a 1--5 scale.}
\label{tab:rationale_quality}
\end{table}
Next, we conduct a human evaluation to assess whether model rationales are useful and grounded, beyond answer accuracy. Annotators are presented the UI, persona, question, answer options, predicted answer, and model explanation, and rate each rationale on UI grounding, persona use, contrastiveness, and overall usefulness. Table~\ref{tab:rationale_quality} shows that SFT produces the weakest explanations, indicating that label supervision alone does not teach the model to justify why an option fits the UI and persona. Models trained with contrastive rationales receive higher ratings, especially for distinguishing the selected answer from plausible alternatives. \textsc{PerceptUI} achieves the highest scores across all criteria, indicating that its rationales are more consistently UI-grounded, persona-aware, and informative about the predicted answer.

\subsection{Persona-Conditioned UI Rating}
\label{sec:exp_reinecke}

\begin{figure*}[tbp]
\centering
\includegraphics[width=0.75\linewidth]{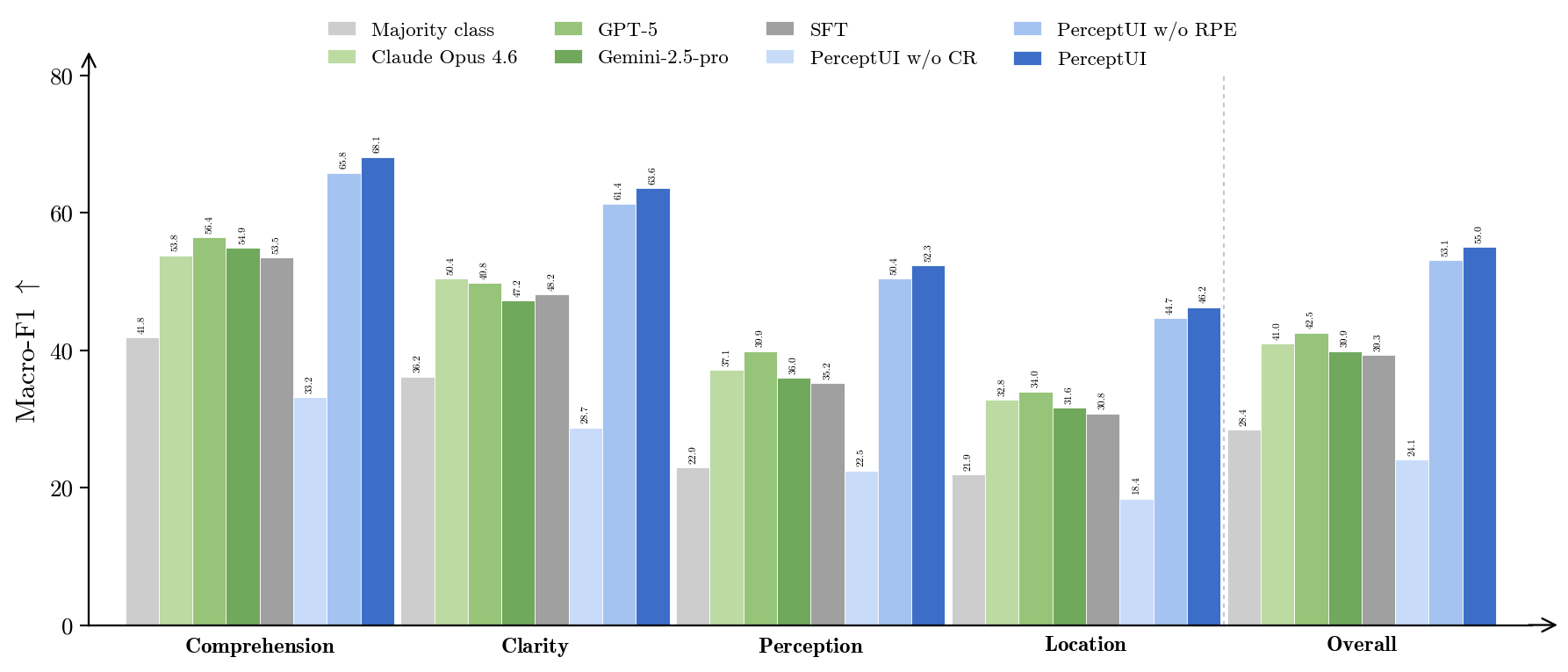}
\caption{UX question answering on UXcar. We report macro-F1 across four groups of questions.}
\label{fig:uxcar_axes}
\end{figure*}

\begin{table}[tbp]
\centering
\small
\resizebox{1.0\linewidth}{!}{
\begin{tabular}{llcccc}
\toprule
\textbf{Method} & \textbf{Setting} & \textbf{Acc.}$\uparrow$ & \textbf{MAE}$\downarrow$ & \textbf{JS Div.}$\downarrow$ & $\boldsymbol{\rho}$$\uparrow$ \\
\midrule
Majority class & - & 24.13 & 1.92 & 0.282 & -- \\
\midrule
GPT-4o & Zero-shot & 27.84 & 1.42 & 0.184 & 0.412 \\
Claude Opus 4.6 & Zero-shot & 28.56 & 1.36 & 0.171 & 0.437 \\
Qwen2.5-VL-32B & Zero-shot & 26.09 & 1.51 & 0.198 & 0.391 \\
GPT-5.5 & Zero-shot & 29.45 & 1.22 & 0.150 & 0.429 \\
\midrule
No persona & SFT & 30.42 & 1.31 & 0.156 & 0.461 \\
Generic persona & SFT & 32.18 & 1.22 & 0.142 & 0.495 \\
\textsc{PerceptUI} w/ generic persona & Generic persona & 35.46 & 1.16 & 0.135 & 0.521 \\
\textsc{PerceptUI} w/o CR & Participant profile & 18.74 & 2.46 & 0.318 & 0.282 \\
\textsc{PerceptUI} w/o RPE 
& Participant profile 
& \cellcolor{wiseryellow}\underline{41.23} 
& \cellcolor{wiseryellow}\underline{0.96} 
& \cellcolor{wiseryellow}\underline{0.103} 
& \cellcolor{wiseryellow}\underline{0.621} \\
\rowcolor{blue!8}
\textsc{PerceptUI} 
& Participant profile 
& \cellcolor{wiserblue}\textbf{43.51} 
& \cellcolor{wiserblue}\textbf{0.88} 
& \cellcolor{wiserblue}\textbf{0.092} 
& \cellcolor{wiserblue}\textbf{0.658} \\
\bottomrule
\end{tabular}}
\caption{UI rating on LabintheWild. Accuracy and $\rho$ are higher better; MAE and JS divergence are lower better.}
\label{tab:reinecke}
\end{table}
Persona-conditioned UI rating is evaluated on the LabintheWild website-aesthetics dataset~\citep{reinecke2014quantifying}, where participants rate web screenshots on a $1$--$9$ Likert scale and each rating is linked to participant profile information. Table~\ref{tab:reinecke} shows that zero-shot MLLMs improve only moderately over the majority class, while supervised baselines benefit from task-specific training. \textsc{PerceptUI} performs best across all metrics, improving exact-match accuracy, reducing rating error, and better matching empirical per-screenshot rating distributions. The gap between no-persona and persona-conditioned variants indicates that participant profiles provide a useful signal for subjective UI judgments.

\subsection{Persona-Conditioned UX Perception}
\label{sec:exp_miniukovich}

\begin{table}[tbp]
\centering
\small
\resizebox{1.0\linewidth}{!}{
\begin{tabular}{llcccc}
\toprule
\textbf{Method} & \textbf{Setting} & \textbf{Acc.}$\uparrow$ & \textbf{MAE}$\downarrow$ & \textbf{JS Div.}$\downarrow$ & $\boldsymbol{\rho}$$\uparrow$ \\
\midrule
Majority class & - & 31.42 & 1.58 & 0.224 & -- \\
\midrule
GPT-4o & Zero-shot & 38.04 & 1.18 & 0.158 & 0.453 \\
Claude Opus 4.6 & Zero-shot & 39.81 & 1.12 & 0.146 & 0.476 \\
Gemini-2.5-pro & Zero-shot & 37.55 & 1.21 & 0.166 & 0.439 \\
GPT-5 & Zero-shot & 38.87 & 1.12 & 0.139 & 0.477 \\
\midrule
No persona & SFT & 42.07 & 1.07 & 0.131 & 0.501 \\
Generic persona & SFT & 44.18 & 1.01 & 0.118 & 0.534 \\
\textsc{PerceptUI} w/ generic persona & Generic persona & 47.62 & 0.93 & 0.108 & 0.568 \\
\textsc{PerceptUI} w/o CR & Participant profile & 10.83 & 2.86 & 0.494 & 0.235 \\
\textsc{PerceptUI} w/o RPE
& Participant profile
& \cellcolor{wiseryellow}\underline{54.16}
& \cellcolor{wiseryellow}\underline{0.79}
& \cellcolor{wiseryellow}\underline{0.083}
& \cellcolor{wiseryellow}\underline{0.661} \\
\rowcolor{blue!8}
\textsc{PerceptUI}
& Participant profile
& \cellcolor{wiserblue}\textbf{56.28} 
& \cellcolor{wiserblue}\textbf{0.71} 
& \cellcolor{wiserblue}\textbf{0.072} 
& \cellcolor{wiserblue}\textbf{0.703} \\
\bottomrule
\end{tabular}}
\caption{UX perception on LabintheWild-UX. }
\label{tab:miniukovich}
\end{table}
LabintheWild-UX extends persona-conditioned rating from visual appeal to broader UX perception. The dataset contains user ratings of UI screenshots across four website domains, with axes covering prototypicality, aesthetics, trustworthiness, and pre-use usability. This experiment evaluates whether \textsc{PerceptUI} can predict multiple aspects of perceived UX. We report metrics averaged across the six rating axes in Table~\ref{tab:miniukovich}. The results show that zero-shot MLLMs remain weaker than supervised models across these UX dimensions. Fine-tuning without persona information improves performance, but persona-aware variants are consistently stronger. \textsc{PerceptUI} achieves the best overall results, reducing rating error and better matching empirical response distributions. This highlights that participant-level context is useful for UX perception, where judgments of familiarity, trustworthiness, and expected usability depend partly on user expectations.

\subsection{UX Question Answering for Automotive Interfaces}
We next examine whether \textsc{PerceptUI} transfers beyond web and mobile interfaces to automotive UX scenarios. UXCar contains participant responses to questions about vehicle-interface screenshots, where answers depend on visual cues, interface conventions, and familiarity with automotive displays. Questions are divided into four groups: comprehension, clarity, perception, and spatial location. Figure~\ref{fig:uxcar_axes} reports macro-F1 for each group. Questions about action and interface clarity are generally easier, as they rely on more explicit screen cues. Perception and spatial location are more challenging as they require attention to specific visual elements and may depend on participants' familiarity with in-car interfaces. \textsc{PerceptUI} achieves the strongest performance across all groups. The drop without contrastive reflection further suggests that explaining why one answer fits better than the alternatives helps the model capture user-specific judgment cues beyond standard answer imitation.

\section{Conclusion}
\label{sec:conclusion}

We introduced \textsc{PerceptUI}, a framework for automated UI/UX evaluation. By combining contrastive reflection fine-tuning with reflective prompt evolution, \textsc{PerceptUI} predicts how a user would answer UI/UX questions and explains the prediction relative to alternative answers. Experiments demonstrate improvements in both personalized and non-personalized UI/UX judgments across diverse tasks, including design selection, rating prediction, population-level response estimation, and critique generation. Additional analyses show that strong performance depends not only on the underlying vision-language model, but also on grounding predictions in user context, contrasting alternative answers, and making the model’s internal evidence explicit through structured explanations. These results highlight \textsc{PerceptUI} as a promising foundation for research and industry applications at the intersection of user modeling, interface evaluation, and human-centered product development.

\section{Limitations}
Despite achieving strong performance across our evaluations, several limitations should be acknowledged. First, the reproducibility of some experiments is limited because our proprietary dataset is not publicly available. Second, \textsc{PerceptUI} may inherit biases from the underlying vision-language model and from the data used to construct personas, including cultural, gender, age, accessibility, and socioeconomic biases. This is especially important because the framework predicts user responses and could otherwise over-represent the preferences of groups that are more visible in the training data.

\noindent Another limitation is that UI/UX evaluation is inherently interactive. Many usability problems only emerge through multi-step interaction, such as navigation breakdowns, delayed feedback, error recovery, form completion, or mismatches between user expectations and system responses. \textsc{PerceptUI} operates on screenshots and evaluation questions, and therefore cannot fully capture the temporal and procedural aspects of user experience. While the WebDevJudge results demonstrate that the framework can transfer to generated web interfaces in an image-only setting, modeling interaction traces, browser actions, and task-completion outcomes remains an important direction for future work.

\noindent In addition, the generated rationales should also be interpreted with care. Although contrastive reflection improves rationale quality, these explanations are model-generated and should not be treated as the true causal reasons behind a participant's response. 

\noindent Another limitation is that the set of baselines varies across experiments. This is because the benchmarks differ substantially in task format, available inputs, and evaluation protocol: some focus on pairwise UI selection, others on rating prediction, explanation recovery, or open-ended critique. As a result, not all prior methods are applicable to every setting, and some benchmarks provide only domain-specific baselines. We therefore compare against the strongest available and compatible baselines for each task, but the results should be viewed as task-specific comparisons rather than a single unified leaderboard across all benchmarks.

\noindent Finally, \textsc{PerceptUI} depends on the quality of the teacher model, the base vision-language model, and the available persona information. Errors in these components can lead to incorrect predictions, overconfident population estimates, or plausible but weakly supported rationales. Our ablations examine these factors, but further work is needed on reproducibility, bias auditing, dynamic interaction modeling, and human validation.

\section{Ethics Statement}
\label{sec:ethics}

This paper introduces an MLLM-based framework for persona-conditioned UI evaluation, enabling early-stage assessment of how different users may perceive and respond to interfaces. While this can support faster and more targeted design iteration, it also raises ethical considerations.

\noindent First, \textsc{PerceptUI} may reproduce or amplify biases present in the underlying vision-language model, the participant data, or the persona representation. Predicted responses may therefore reflect stereotypes about age, gender, culture, disability, digital literacy, occupation, or socioeconomic status rather than the views of real users. This is particularly important when the system is used to compare designs for different user groups, since biased predictions could lead to interface decisions that benefit some groups while disadvantaging others.

\noindent Second, simulated user responses should not be treated as a replacement for human participation. Although synthetic evaluation can reduce the cost of early exploration, relying on it too heavily may marginalize the voices of actual users, especially users from underrepresented or accessibility-sensitive groups. There is also a risk that user-conditioned predictions could be used to optimize interfaces for persuasion or behavioral manipulation rather than usability, transparency, or user well-being. Such applications require careful oversight, clear documentation, and human review.

\noindent Finally, participant-linked UI data can contain sensitive information about preferences, habits, accessibility needs, or decision-making patterns. Any deployment of \textsc{PerceptUI} should therefore follow appropriate privacy safeguards, limit unnecessary data retention, and avoid using inferred personas to make consequential decisions about individuals. We view \textsc{PerceptUI} as a tool to complement, not replace, human-centered design practices. By using synthetic evaluation as an early diagnostic signal while preserving human validation in product decisions, we aim to support UI evaluation that is transparent, inclusive, and socially responsible.

\bibliography{custom}

\clearpage
\appendix

\section{Experimental Setup}

\begin{figure}[tbp]
    \centering
    \includegraphics[width=1.0\linewidth]{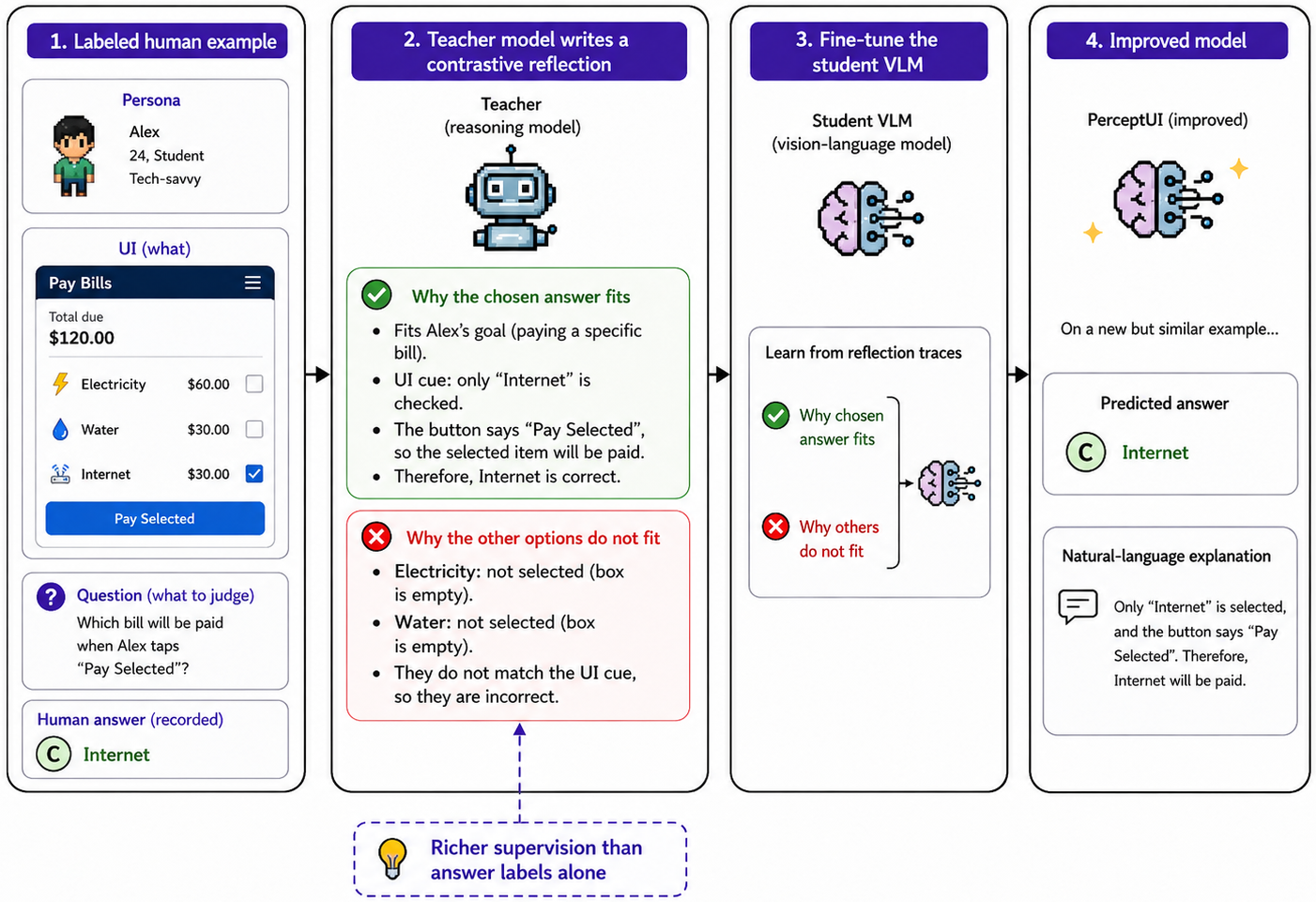}
    \caption{Contrastive reflection fine-tuning.}
    \label{fig:figure2}
\end{figure}

Different benchmarks provide different forms of supervision. We convert all benchmarks into the same input-output interface used by \textsc{PerceptUI}: a UI screenshot or screenshot pair, an evaluation question, a set of answer options, and an optional persona. When participant-level profiles are available, they are included as persona context. When no participant profile is provided, the persona field is omitted, and the model is evaluated as a non-personalized UI evaluator. For pairwise preference benchmarks, the answer options correspond to the two UI variants. For rating benchmarks, the answer options correspond to the discrete rating scale. For explanation benchmarks, the model is prompted to generate a rationale or critique conditioned on the benchmark-specific target format. Unless otherwise stated, we use GPT-5.5 as the teacher model for generating contrastive rationales during contrastive reflection fine-tuning (Figure \ref{fig:figure2}).

To avoid leakage between model training and prompt selection, we use disjoint data for contrastive reflection fine-tuning and reflective prompt evolution. For each dataset, examples are first split into non-overlapping training, development, and test partitions. The training partition is used only for contrastive reflection fine-tuning: teacher rationales, question paraphrases, UI grounding summaries, and persona grounding summaries are generated only for training examples, and only these examples are used to update the student parameters $\theta$. The development samples are used only for reflective prompt evolution. During this stage, $\theta$ is fixed, and no gradient update is performed. Development examples are used to evaluate prompt candidates, compute symbolic feedback, and select the final prompt $P^\star$, but they are never used to fine-tune the model or to generate supervised training targets for CRFT. The test partition is held out from both stages and is used only for final evaluation. This separation ensures that the model is not fine-tuned on the same examples used to optimize prompts, and that reported test performance does not rely on examples seen during either parameter training or prompt evolution.
\begin{figure}[tbp]
    \centering
    \includegraphics[width=1.0\linewidth]{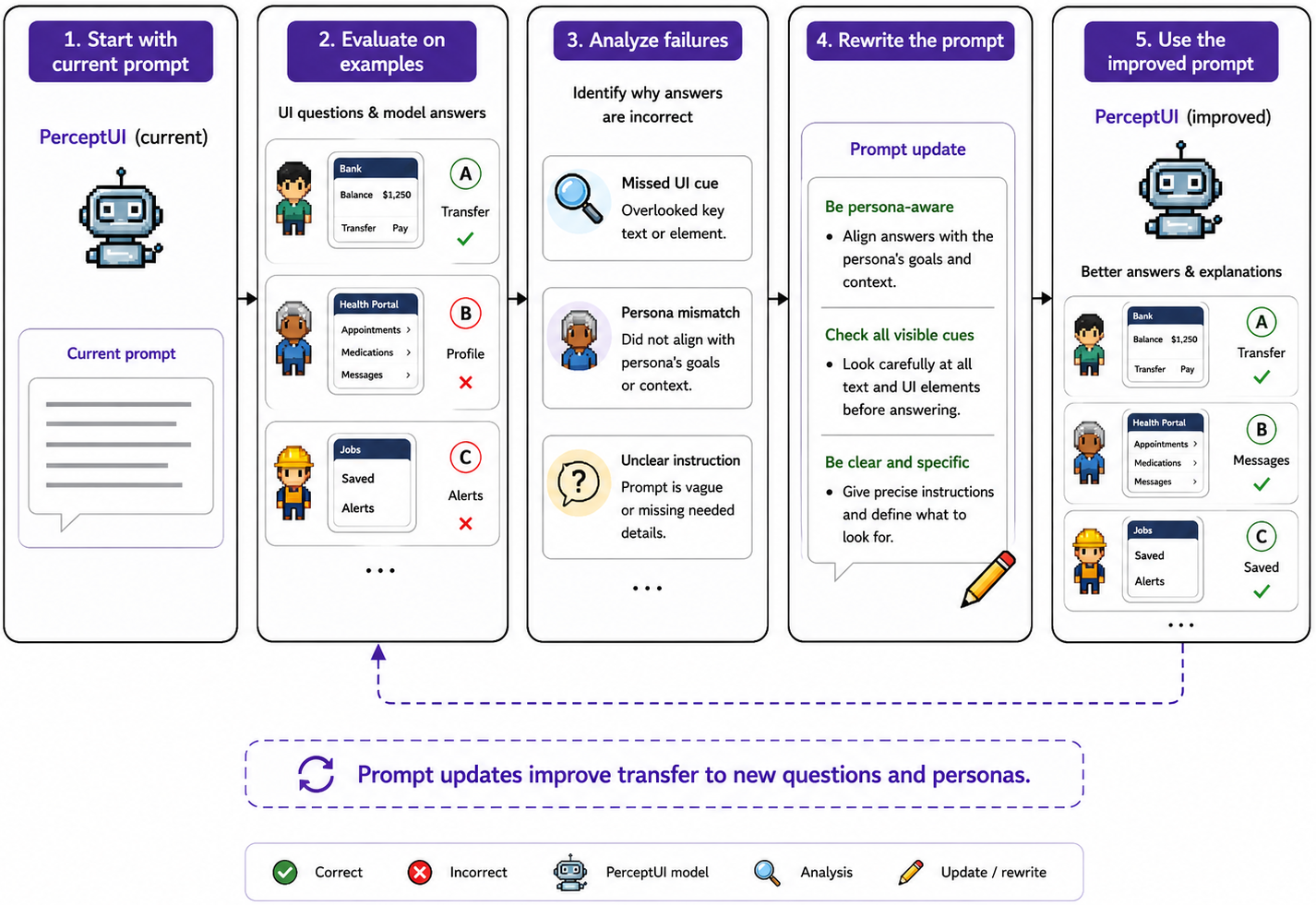}
    \caption{Reflective prompt evolution.}
    \label{fig:figure3}
\end{figure}

After each prompt update during prompt evolution (see Figure \ref{fig:figure3}), we run an additional LLM-based validation step before adding the candidate prompt to the pool. The judge verifies that the revised prompt does not contain gold labels, empirical answer distributions, dataset-specific shortcuts, example identifiers, or rules that directly encode development-set failures. It also checks that the update does not introduce overly specific instructions tied to particular screenshots, personas, questions, or answer options. Prompts that fail this check are rejected. In addition, the prompt optimizer is explicitly instructed to avoid leakage, memorization, dataset-specific heuristics, and narrow rules. Instead, it must propose general edits to the decision procedure, such as improving the use of UI evidence, answer-option comparison, or persona-question grounding.

\begin{table}[tbp]
\centering
\small
\resizebox{\columnwidth}{!}{
\begin{tabular}{ll}
\toprule
\textbf{Component} & \textbf{Setting} \\
\midrule
Student model & Qwen3-VL-8B-Instruct \\
Fine-tuning method & QLoRA (4-bit, $r{=}16$, $\alpha{=}32$, dropout $0.05$) \\
Optimizer & AdamW \\
Precision / memory & bf16, gradient checkpointing \\
Learning rate & $2{\times}10^{-4}$, warm-up ratio $0.1$ \\
Batch size & $1$ per device, gradient accumulation $2$ \\
Training epochs & $3$ \\
Image pixel budget & $\min{=}256{\cdot}28^2$, $\max{=}512{\cdot}28^2$ pixels \\
Teacher model for CRFT & GPT-5.5 \\
RPE rounds $T$ & $24$ \\
RPE minibatch size & $4$ \\
Prompt candidates per round & $6$ \\
Question paraphrases $K$ & $3$ \\
Evaluator / analyzer / optimizer & GPT-5.5 \\
Evaluation decoding & temperature $0$, max tokens $4,096$ \\
Random seeds & $\{42, 7, 123\}$ \\
\bottomrule
\end{tabular}}
\caption{Implementation details for contrastive reflection fine-tuning and reflective prompt evolution.}
\label{tab:implementation_details}
\end{table}
To make the training and prompt-optimization procedure reproducible, we summarize the main implementation settings in Table \ref{tab:implementation_details}. These include the student model, fine-tuning configuration, teacher-generation settings, reflective prompt evolution parameters, and other relevant parameters.

For LabintheWild and LabintheWild-UX datasets (Tab.~\ref{tab:reinecke}, Tab.~\ref{tab:miniukovich}), we evaluate both individual rating prediction and aggregate alignment. Accuracy measures exact agreement with the recorded rating, while MAE measures the average distance from the target rating and therefore accounts for near misses on fine-grained scales. JS divergence compares predicted and empirical rating distributions for each screenshot, and Spearman $\rho$ measures whether the model ranks interfaces in the same order as human ratings. Since the majority-class baseline assigns the same prediction to all examples, its rank correlation is undefined. For the UXcar analysis (Fig.~\ref{fig:uxcar_axes}), each agent predicts the answer of a matched participant. We therefore report macro-F1, which averages performance across answer options and reduces the influence of dominant responses. The overall score is computed as the mean across questions and corresponds to the macro-F1 reported in Tab.~\ref{tab:ablation}. We report reasoning-enhanced prompting baselines, including CoCoT, Self-Refine, DDCoT, and multi-agent debate, on WiserUI-Bench following its evaluation protocol. We do not apply these baselines to all pairwise-selection benchmarks because their relevance depends on the task format and metrics. In particular, multi-agent debate is mainly designed to reduce order-dependent errors, which WiserUI-Bench captures through FA, SA, and Consistent Accuracy. UIClip/BetterApp, in contrast, reports a single overall-accuracy metric and focuses on visual design-quality preference rather than A/B-test outcome prediction. Its dominant challenge is therefore perceptual grounding of design quality, rather than inconsistent reasoning across input orders. We consequently report reasoning-enhanced prompting baselines where they are diagnostically informative, and use the strongest task-appropriate baselines for the remaining benchmarks.

For the human evaluation in Section~\ref{sec:exp_rationale_quality}, we sample 120 UI--persona--question instances from the test split. For each instance, annotators are shown the UI screenshot, persona, question, answer options, predicted answer, and one anonymized rationale. Each rationale is evaluated by three annotators along four dimensions: UI grounding, persona use, contrastiveness, and overall usefulness. Ratings are collected on a 1--5 Likert scale. Model identities are hidden, and rationale order is randomized. We report mean scores across annotators and examples.

\section{Discussion}
\label{sec:discussion}

Experimental results suggest that persona-conditioned UI/UX evaluation offers a useful alternative to treating MLLMs as generic interface judges. Across tasks, \textsc{PerceptUI} is strongest when the model must connect visual evidence, answer options, and user-specific context. This is consistent with the motivation for contrastive reflection: subjective UI judgments often depend not only on why one answer is plausible, but also on why nearby alternatives are less appropriate for a given user.

At the same time, \textsc{PerceptUI} should be viewed as an early-stage evaluation tool rather than a replacement for human studies or online A/B tests. Its predictions can help identify likely usability issues, compare design alternatives, and prioritize interfaces for further testing, but final product decisions should still require validation with real users. This distinction is important because the model can approximate patterns in recorded responses, but it cannot fully observe the motivations, expectations, or situational factors that shape an individual user's judgment.

Our experiments also highlight the importance of persona quality. Generic or mismatched personas provide limited benefit, while participant-level information improves prediction and calibration. This suggests that persona-conditioned evaluation depends not only on whether user information is included, but also on whether the representation captures factors relevant to the UI question. Richer personas may improve fidelity, but they also require stronger privacy safeguards and more careful validation.

An important implication of our results is that individual-level prediction and population-level calibration should be interpreted separately. \textsc{PerceptUI} may still make mistakes for a particular user, especially when the available persona omits relevant context. However, aggregating predictions across many profiles can produce useful estimates of population-level response distributions. This makes the framework more appropriate for early design screening, subgroup analysis, and prioritizing follow-up studies than for making definitive claims about any single user. 

Finally, the available benchmarks considered in this work cover complementary but incomplete views of UI/UX quality, including preference prediction, design quality, generated web interfaces, critique quality, and UX perception. Future work should extend UI/UX evaluation to interaction with the interface and longer-term user histories, so that models can better capture how users perceive, navigate, and adapt to interfaces over time.

\section{Cost Analysis}
\label{sec:cost_analysis}
We report the cost of \textsc{PerceptUI} for evaluating one UI with five questions and 20 personas. Frontier models tend to be inexpensive for a single judgment, but become costly when each UI must be evaluated separately for multiple personas. \textsc{PerceptUI} shifts part of the computation to an offline stage, where contrastive rationales are generated, and the student model is fine-tuned. Once trained, UI evaluation can be performed with a smaller hosted model or a self-hosted VLM. Table~\ref{tab:cost_analysis} shows that frontier-model evaluation grows linearly with the number of sampled personas, since each persona requires a separate query. In contrast, \textsc{PerceptUI} incurs a one-time preparation cost but substantially reduces the per-UI inference cost. This makes large-scale evaluation of UI variants more practical than repeatedly querying a frontier model for each simulated user.

\begin{table}[tbp]
\centering
\small
\begin{tabular}{lc}
\toprule
\textbf{Method} & \textbf{Cost} \\
\midrule
Frontier MLLM, no personas & \$0.05--\$0.06 / UI \\
Frontier MLLM, 20 personas & \$1.05--\$1.20 / UI \\
Low-cost MLLM, 20 personas & \$0.04--\$0.10 / UI \\
\midrule
\textsc{PerceptUI} reflection generation & \$150--\$300 one-time \\
\textsc{PerceptUI} fine-tuning & \$50--\$200 one-time \\
\rowcolor{blue!8}
\textsc{PerceptUI} inference & \$0.07--\$0.25 / UI \\
\bottomrule
\end{tabular}
\caption{Cost for evaluating one UI with five questions and 20 personas.}
\label{tab:cost_analysis}
\end{table}

\section{Prompts}
\label{appendix:prompts}
We provide simplified prompt templates used by \textsc{PerceptUI}. The full prompts include dataset-specific formatting instructions and output constraints, which are omitted for readability.

\subsection{UI Grounding Prompt}
\label{appendix:ui-grounding}

The UI grounding prompt produces the visual summary $u_i$ used by the answer simulator and, during training, by the contrastive reflection teacher.

\begin{tcolorbox}[colframe=customblues3, colback=white, title=UI Grounding Prompt, breakable]
Analyze the UI screenshot with respect to the given UX question. Focus only on visual evidence that is visible in the image or reference image. Do not infer information that is not shown.\\

\textbf{UI image:} \textcolor{customorange}{\texttt{\{ui\_image\}}}\\
\textbf{Reference image (optional):} \textcolor{customorange}{\texttt{\{reference\_image\}}}\\
\textbf{Question:} \textcolor{customorange}{\texttt{\{question\_text\}}}\\
\textbf{Options:} \textcolor{customorange}{\texttt{\{response\_options\}}}\\

Output the following:
\begin{itemize}
  \item \textbf{Question-relevant UI evidence:} visible elements that may affect the answer.
  \item \textbf{Visual hierarchy:} what the user is likely to notice first and how attention may flow across the screen.
  \item \textbf{Potential ambiguity:} UI elements that may be unclear, visually subtle, or open to multiple interpretations.
\end{itemize}
\end{tcolorbox}

\subsection{Persona Grounding Prompt}
\label{appendix:persona-grounding}

The persona grounding prompt produces the persona summary $g_i$ used by the answer simulator and contrastive reflection teacher.

\begin{tcolorbox}[colframe=customblues3, colback=white, title=Persona Grounding Prompt, breakable]
Given a participant description, identify which aspects of the persona may affect how this participant answers the UX question. Use only information stated in the persona. Do not introduce stereotypes or assumptions beyond the provided description.\\

\textbf{Persona:} \textcolor{customorange}{\texttt{{persona\_description}}}\\
\textbf{Question:} \textcolor{customorange}{\texttt{{question\_text}}}\\
\textbf{Options:} \textcolor{customorange}{\texttt{{response\_options}}}\\
\textbf{UI grounding summary:} \textcolor{customorange}{\texttt{{ui\_summary}}}\\

Output the following:
\begin{itemize}
\item \textbf{Relevant persona factors:} preferences, constraints, familiarity, goals, or sensitivities that may affect the response.
\item \textbf{Likely interpretation:} how these factors may influence what the participant notices or values in the UI.
\item \textbf{Irrelevant factors:} persona details that should not affect this specific question.
\end{itemize}
\end{tcolorbox}

\subsection{Persona-Conditioned Answer Simulation}
\label{appendix:answer-simulation}

At inference, the answer simulator predicts a rationale $\hat{z}_i$ and an answer $\hat{y}_i$ from the UI grounding summary $u_i$, persona grounding summary $g_i$, question, and answer options.

\noindent\textbf{Multiple-choice questions.}
\begin{tcolorbox}[colframe=customblues3, colback=white, title=Multiple-Choice Answer Simulation Prompt, breakable]
Simulate how the described participant would answer the UX question. Base the answer on the UI screenshot, UI evidence, persona grounding summary, and answer options. If persona information is not relevant, rely primarily on the visible UI evidence.\\

\textbf{UI image:} \textcolor{customorange}{{ui\_image}}\\
\textbf{Reference image (optional):} \textcolor{customorange}{{reference\_image}}\\
\textbf{Persona:} \textcolor{customorange}{{persona\_description}}\\
\textbf{UI grounding summary:} \textcolor{customorange}{{ui\_summary}}\\
\textbf{Persona grounding summary:} \textcolor{customorange}{{persona\_summary}}\\
\textbf{Question:} \textcolor{customorange}{{question\_text}}\\
\textbf{Options:} \textcolor{customorange}{{response\_options}}\\
\textbf{Reference description (optional):} \textcolor{customorange}{{reference\_description}}\\

Output the following:
\begin{itemize}
\item \textbf{Rationale:} a brief explanation of why the selected option best matches the UI evidence and persona.
\item \textbf{Answer:} one option from the provided answer set.
\item \textbf{Confidence:} \textit{very\_confident}, \textit{somewhat\_confident}, \textit{uncertain}, or \textit{guessing}.
\end{itemize}
\end{tcolorbox}

\noindent\textbf{Location-grid questions.}

\begin{tcolorbox}[colframe=customblues3, colback=white, title=Location-Grid Answer Simulation Prompt, breakable]
Simulate how the described participant would answer a spatial question on a UI divided into grid areas. The grid options are A--G, with H indicating ``Not sure.'' Use visible UI evidence and the persona grounding summary. Do not assume that the participant sees elements that are not visible in the screenshot.\\

\textbf{UI image with grid overlay:} \textcolor{customorange}{{ui\_image}}\\
\textbf{Persona:} \textcolor{customorange}{{persona\_description}}\\
\textbf{UI grounding summary:} \textcolor{customorange}{{ui\_summary}}\\
\textbf{Persona grounding summary:} \textcolor{customorange}{{persona\_summary}}\\
\textbf{Question:} \textcolor{customorange}{{question\_text}}\\
\textbf{Grid options:} \textcolor{customorange}{{grid\_options}}\\
\textbf{Reference image (optional):} \textcolor{customorange}{{reference\_image}}\\

Output the following:
\begin{itemize}
\item \textbf{Rationale:} a brief explanation of which grid areas are plausible and why.
\item \textbf{Answer:} one option from A--H.
\item \textbf{Confidence distribution:} formatted as \texttt{A,B,...,H}, summing to 100.
\end{itemize}
\end{tcolorbox}

\noindent\textbf{Likert-scale questions.}

\begin{tcolorbox}[colframe=customblues3, colback=white, title=Likert-Scale Answer Simulation Prompt, breakable]
Simulate how the described participant would answer a Likert-scale UI/UX question. Use the provided scale anchors exactly, and select the rating that best reflects the participant's likely response to the UI.\\

\textbf{UI image:} \textcolor{customorange}{{ui\_image}}\\
\textbf{Reference image (optional):} \textcolor{customorange}{{reference\_image}}\\
\textbf{Persona:} \textcolor{customorange}{{persona\_description}}\\
\textbf{UI grounding summary:} \textcolor{customorange}{{ui\_summary}}\\
\textbf{Persona grounding summary:} \textcolor{customorange}{{persona\_summary}}\\
\textbf{Question:} \textcolor{customorange}{{question\_text}}\\
\textbf{What is rated:} \textcolor{customorange}{{rated\_item}}\\
\textbf{Scale anchors:} \textcolor{customorange}{{scale\_anchors}}\\
\textbf{Reference image (optional):} \textcolor{customorange}{{reference\_image}}\\

Output the following:
\begin{itemize}
\item \textbf{Rationale:} a brief explanation grounded in the UI evidence and persona summary.
\item \textbf{Rating:} an integer within the given scale.
\item \textbf{Confidence:} \textit{very\_confident}, \textit{somewhat\_confident}, \textit{uncertain}, or \textit{guessing}.
\end{itemize}
\end{tcolorbox}

\subsection{Contrastive Reflection Teacher}
\label{appendix:crft-teacher}

For each training example $(x_i, y_i^\star)$, the teacher $T$ is prompted with the inputs of $x_i$, the UI grounding summary $u_i$, the persona grounding summary $g_i$, and the recorded human answer $y_i^\star$. The teacher produces the contrastive reflection $c_i^\star$ used as the CRFT supervisory target.

\begin{tcolorbox}[colframe=customblues3, colback=white, title=Contrastive Reflection Teacher Prompt, breakable]
You will be given a UI screenshot, an optional reference image, a UX question, answer options, a participant description, UI and persona grounding summaries, and the answer this participant actually selected. Write a short contrastive reflection explaining why the recorded answer fits this participant better than the alternatives.\\

\textbf{UI image:} \textcolor{customorange}{{ui\_image}}\\
\textbf{Reference image (optional):} \textcolor{customorange}{{reference\_image}}\\
\textbf{Question:} \textcolor{customorange}{{question\_text}}\\
\textbf{Options:} \textcolor{customorange}{{response\_options}}\\
\textbf{Participant:} \textcolor{customorange}{{persona\_description}}\\
\textbf{UI grounding summary:} \textcolor{customorange}{{ui\_summary}}\\
\textbf{Persona grounding summary:} \textcolor{customorange}{{persona\_summary}}\\
\textbf{Recorded answer:} \textcolor{customorange}{{gold\_answer}}\\

Write the reflection in three parts:
\begin{itemize}
\item \textbf{UI evidence.} Identify the visible UI evidence relevant to the answer.
\item \textbf{Persona relevance.} Explain which aspects of the participant profile matter for this question, if any.
\item \textbf{Option contrast.} Explain why the recorded answer fits better than each alternative option.
\end{itemize}

Constraints:
\begin{itemize}
\item Reason from the participant's perspective, not from a model's perspective.
\item Do not state that the recorded answer is ``correct'' in absolute terms; explain why it fits this participant.
\item Do not introduce facts that are not visible in the UI or stated in the participant description.
\item Do not reveal dataset statistics, label frequencies, or any information outside the current example.
\end{itemize}
\end{tcolorbox}

\subsection{Reflective Prompt Evolution}
\label{appendix:rpe}

Reflective prompt evolution updates the prompt configuration $P=(P_{\mathrm{ui}},P_{\mathrm{persona}},P_{\mathrm{ans}})$ while keeping the model parameters fixed. We provide simplified versions of the evaluator, analyzer, optimizer, and audit prompts.

\noindent\textbf{Failure evaluator.}

\begin{tcolorbox}[colframe=customblues3, colback=white, title=Failure Evaluator Prompt, breakable]
You are evaluating a prompt used for persona-conditioned UI/UX answer prediction. You will be given the current prompt, a minibatch of development examples, model predictions, human targets, and, when available, empirical answer distributions. Identify where the prompt fails.\\

\textbf{Current prompt:} \textcolor{customorange}{{current\_prompt}}\\
\textbf{UI images:} \textcolor{customorange}{{ui\_images}}\\
\textbf{Development examples:} \textcolor{customorange}{{dev\_examples}}\\
\textbf{Model outputs:} \textcolor{customorange}{{model\_outputs}}\\
\textbf{Human targets:} \textcolor{customorange}{{human\_targets}}\\
\textbf{Empirical answer distributions (optional):} \textcolor{customorange}{{human\_distributions}}\\

Output the following:
\begin{itemize}
\item \textbf{Residual summary:} where predicted answers or distributions differ from human responses.
\item \textbf{Representative failures:} examples that illustrate recurring errors.
\item \textbf{Likely failure causes:} whether errors are related to UI grounding, persona grounding, answer-option interpretation, scale interpretation, or overuse of generic priors.
\end{itemize}

Do not propose edits yet. Only diagnose failures.
\end{tcolorbox}

\noindent\textbf{Prompt analyzer.}

\begin{tcolorbox}[colframe=customblues3, colback=white, title=Prompt Analyzer Prompt, breakable]
You are given a prompt and a symbolic failure summary. Identify prompt-level weaknesses that may explain the observed errors. Focus on changes that improve general behavior, not example-specific fixes.\\

\textbf{Current prompt:} \textcolor{customorange}{{current\_prompt}}\\
\textbf{Failure summary:} \textcolor{customorange}{{failure\_summary}}\\

Output the following:
\begin{itemize}
\item \textbf{Weak prompt components:} which parts of the prompt likely caused the failures.
\item \textbf{Symbolic gradient:} concise edit instructions for improving the prompt.
\item \textbf{Targeted scope:} whether each edit applies to UI grounding, persona grounding, answer simulation, or output formatting.
\end{itemize}

Constraints:
\begin{itemize}
\item Do not add rules that depend on specific examples, labels, datasets, or answer frequencies.
\item Do not include the correct answer to any development example.
\item Prefer general instructions that improve UI evidence use, persona relevance, and answer-option comparison.
\end{itemize}
\end{tcolorbox}

\noindent\textbf{Prompt optimizer.}

\begin{tcolorbox}[colframe=customblues3, colback=white, title=Prompt Optimizer Prompt, breakable]
Revise the current prompt using the symbolic gradient. Modify only the targeted components and preserve the original task, inputs, and output format unless the symbolic gradient explicitly requests a formatting change.\\

\textbf{Current prompt:} \textcolor{customorange}{{current\_prompt}}\\
\textbf{Symbolic gradient:} \textcolor{customorange}{{symbolic\_gradient}}\\

Output the revised prompt.

Constraints:
\begin{itemize}
\item Preserve the same input variables.
\item Do not add dataset-specific shortcuts or hard-coded decision rules.
\item Do not include development-set labels, answer frequencies, or example-specific hints.
\item Keep the prompt concise enough for repeated inference.
\end{itemize}
\end{tcolorbox}

\noindent\textbf{Prompt audit.}

\begin{tcolorbox}[colframe=customblues3, colback=white, title=Prompt Audit Prompt, breakable]
Audit the candidate prompt before it is added to the prompt pool. Determine whether the prompt is safe to evaluate on held-out data.\\

\textbf{Candidate prompt:} \textcolor{customorange}{{candidate\_prompt}}\\
\textbf{Original prompt:} \textcolor{customorange}{{original\_prompt}}\\
\textbf{Audit checklist:}
\begin{itemize}
\item Does the prompt contain answer leakage?
\item Does the prompt mention development-set labels, examples, or label frequencies?
\item Does the prompt introduce dataset-specific shortcuts?
\item Does the prompt contain hard-coded decision rules?
\item Does the prompt encourage unsupported assumptions about personas or demographics?
\item Does the prompt preserve the intended input variables and output format?
\end{itemize}

Output:
\begin{itemize}
\item \textbf{Decision:} \texttt{accept} or \texttt{reject}.
\item \textbf{Reason:} a brief explanation of the decision.
\item \textbf{Required fix:} if rejected, describe the minimal change needed.
\end{itemize}
\end{tcolorbox}

\section{Datasets}

\begin{table}[tbp]
\centering
\small
\resizebox{1.0\linewidth}{!}{
\begin{tabular}{lc}
\toprule
\textbf{Dataset} & \textbf{Persona Available} \\
\midrule
UXCar & \checkmark \\
WiserUI-Bench~\citep{jeon2025wiserui} & \ding{55} \\
UIClip/BetterApp~\citep{wu2024uiclip} & \ding{55} \\
WebDevJudge~\citep{luera2025webdevjudge} & \ding{55} \\
UICrit~\citep{duan2024uicrit} & \ding{55} \\
LabintheWild ~\citep{reinecke2014quantifying} & \checkmark \\
LabintheWild-UX ~\citep{miniukovich2024prototypicality} & \checkmark \\
\bottomrule
\end{tabular}}
\caption{Persona availability across evaluation datasets.}
\label{tab:persona_availability}
\end{table}
Table~\ref{tab:persona_availability} summarizes whether each dataset provides participant-level profile information. When profiles are available, we include them as persona context. When profiles are unavailable, we omit the persona field and evaluate \textsc{PerceptUI} as a non-personalized UI evaluator.

\paragraph{UXCar Dataset.} We leverage a proprietary participant-level UI/UX evaluation dataset, denoted as \textsc{UXCar}. This benchmark focuses on in-vehicle interface evaluation and contains approximately $500$ participants and $30$ UI/UX questions. Each example consists of a UI screenshot, a participant profile, an evaluation question, and a multiple-choice answer. Participant profiles contain user attributes such as age and gender. Questions cover several aspects of UI/UX perception, including interface understanding, spatial preference, icon interpretation, and perceived usability. Some questions additionally provide a reference icon or auxiliary visual cue together with the main interface screenshot. For example, one question presents a navigation or map interface and asks whether the participant correctly understands a driving instruction, such as where or when to turn. The dataset is designed to capture subjective user judgments conditioned on both visual interface content and participant characteristics, enabling evaluation of persona-conditioned UI response prediction and population-level calibration.

\section{Additional Experiments}

\subsection{Rationale Alignment in UI/UX Selection}
\label{sec:exp_wiserui_interpretation}

\begin{table}[tbp]
\centering
\small
\resizebox{0.95\linewidth}{!}{
\begin{tabular}{lcc}
\toprule
\textbf{Model} & \textbf{Interp. Recall}$\uparrow$ & \textbf{Inst. Recall}$\uparrow$ \\
\midrule
o1 & 64.18 & 78.33 \\
GPT-4o & 50.15 & 66.67 \\
GPT-5 & 53.78 & 69.33 \\
Claude Opus 4.6 & 52.11 & 67.99 \\
Qwen2.5-VL-32B & 50.23 & 63.00 \\
Qwen2.5-VL-7B & 43.71 & 61.67 \\
InternVL-2.5-38B & 45.32 & 59.33 \\
InternVL-2.5-8B & 35.09 & 51.67 \\
LLaVA-NeXT-7B & 11.99 & 21.67 \\
LLaVA-OneVision-7B & 21.78 & 33.33 \\
\midrule
\textsc{PerceptUI} w/o CR & 47.86 & 68.22 \\
\textsc{PerceptUI} w/o RPE & \cellcolor{wiseryellow}\underline{66.42} & \cellcolor{wiseryellow}\underline{80.11} \\
\rowcolor{blue!8}
\textsc{PerceptUI} 
& \cellcolor{wiserblue}\textbf{69.05} 
& \cellcolor{wiserblue}\textbf{84.09} \\
\bottomrule
\end{tabular}}
\caption{Rationale alignment on WiserUI-Bench. Interp. Recall measures recovery of expert-written interpretations, while Inst. Recall measures whether at least one expert interpretation is recovered for each UI pair.}
\label{tab:wiserui_interpretation}
\end{table}
WiserUI-Bench also provides expert-written rationales for the A/B-test winner in each UI pair. This allows us to test whether model explanations recover the design factors experts identify as supporting the more effective variant. Following the benchmark protocol, the winning UI is given, and the model generates an explanation for the preference. As shown in Table~\ref{tab:wiserui_interpretation}, prior methods recover only part of the expert rationale space. \textsc{PerceptUI} achieves the highest interpretation-level and instance-level recall, indicating stronger alignment with human UI/UX interpretations. These results further show that the agent can associate design preferences with concrete UI factors.

\subsection{Quality of UI Critiques}
\label{sec:exp_uicrit_critique}

\begin{table}[tbp]
\centering
\small
\resizebox{0.92\linewidth}{!}{
\begin{tabular}{lccc}
\toprule
\textbf{Method} & \textbf{Comment Quality}$\uparrow$ & \textbf{\# Comments} & \textbf{Avg. Rank}$\downarrow$ \\
\midrule
Zero-shot Gemini & 0.31 & 29 & 6.2 \\
Zero-shot GPT-5 & 0.37 & 22 & 5.1 \\
Zero-shot Claude Opus 4.6 & 0.35 & 24 & 4.8 \\
UICrit few-shot + visual prompting & 0.48 & 14 & 4.2 \\
Human designer critiques
& \cellcolor{wiserblue}\textbf{0.75}
& 17
& \cellcolor{wiserblue}\textbf{1.5} \\
\midrule
\textsc{PerceptUI} w/o CR & 0.27 & 20 & 7.7 \\
\textsc{PerceptUI} w/o RPE & 0.52 & 15 & 3.8 \\
\rowcolor{blue!8}
\textsc{PerceptUI}
& \cellcolor{wiseryellow}\underline{0.54}
& 16
& \cellcolor{wiseryellow}\underline{2.7} \\
\bottomrule
\end{tabular}}
\caption{UI critique quality on UICrit, comparing automatic feedback with human designer critiques.}
\label{tab:uicrit_critique}
\end{table}
Beyond selecting preferred designs or identifying relevant design principles, UI/UX evaluation also requires actionable feedback that tells designers what to improve and where. UICrit~\citep{duan2024uicrit} provides a suitable benchmark for evaluating UI critique quality, as it contains expert-written critiques for mobile UI screenshots, with comments linked to relevant screen regions and design-quality ratings. We compare automatic feedback against zero-shot Gemini, the strongest few-shot visual-prompting baseline, and human designer critiques. Table \ref{tab:uicrit_critique} suggests that human designer critiques remain stronger than all LLM methods, highlighting the difficulty of open-ended UI feedback. It further indicates that our model produces more useful critiques than other baselines, despite using a substantially smaller model than frontier proprietary MLLMs. This suggests that few-shot imitation of critique examples is insufficient for producing grounded feedback, often leading to shallow comments that do not capture why a design issue matters.

\subsection{Web Interface Preference}
\label{sec:exp_webdevjudge}

\begin{figure}[tbp]
\centering
\includegraphics[width=1.0\linewidth]{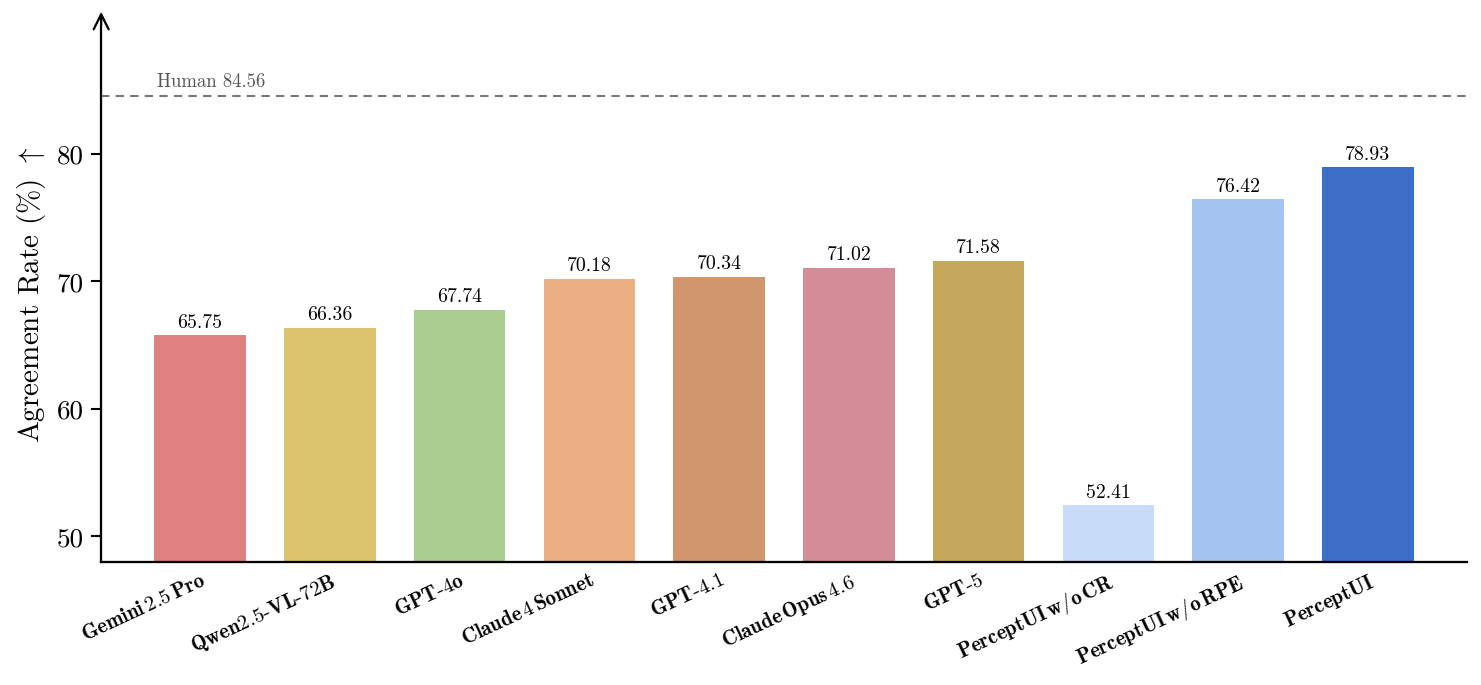}
\caption{Interface preference on WebDevJudge. The model receives two screenshots of web implementations generated from the same user request and selects the preferred design.}
\label{fig:webdevjudge_image}
\end{figure}

With the rise of AI-generated interfaces, UX evaluation increasingly involves comparing alternative web prototypes produced from the same user request. We therefore evaluate on WebDevJudge~\citep{luera2025webdevjudge}, which measures preferences between generated web implementations. This setting is less standardized than WiserUI-Bench or UIClip as interfaces may differ not only in visual design, but also in layout, content organization, and how well they satisfy the requested task. Given two UI screenshots, the model is prompted to select the preferred design. Figure \ref{fig:webdevjudge_image} reveals that agreement remains low for all image-only judges, with the strongest baselines differing by only a few points. \textsc{PerceptUI} obtains the best agreement, suggesting that its UI judgment transfers to less standardized, generated interfaces. The modest improvement further indicates that some preferences depend on task satisfaction, interaction flow, or implementation details that cannot be fully inferred from a static screenshot.

\subsection{UI Preference Explanation}
\label{sec:exp_uiclip_suggestion}

\begin{figure}[tbp]
\centering
\includegraphics[width=1.0\linewidth]{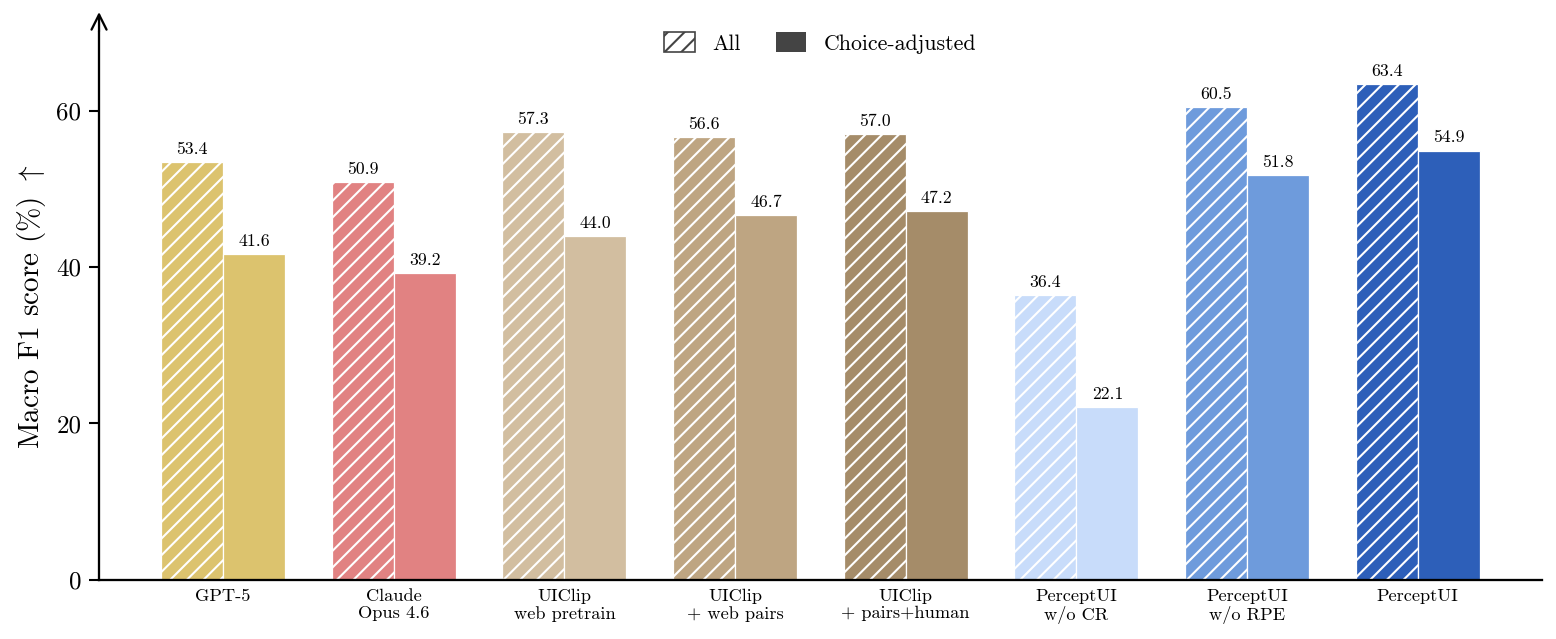}
\caption{UI preference explanation on UIClip~\citep{wu2024uiclip}. Given a pair of UI screenshots and the preferred design, it predicts CRAP principles.}
\label{fig:uiclip_suggestion}
\end{figure}

In practice, UX researchers and designers need more than a preference label. They also need to know which design principles explain why one interface is better than another. We assess this task on UIClip/BetterApp~\citep{wu2024uiclip}. Given a pair of UI screenshots and the preferred design, the agent predicts which CRAP principles (contrast, repetition, alignment, and proximity) explain the preference. This task measures whether the model can connect design-relevant evidence behind a choice, rather than simply selecting a higher-quality UI. Following UIClip, we report the macro-averaged F1 over the four CRAP principles, together with a choice-adjusted F1 that ignores a model's suggestions when it selects the wrong preferred UI. Figure ~\ref{fig:uiclip_suggestion} illustrates that frontier MLLMs achieve competitive F1 but much lower choice-adjusted F1, consistent with their tendency to mention many plausible principles even when the selected UI is incorrect. \textsc{PerceptUI} achieves the strongest F1 under both metrics. This suggests that its suggestions are better tied to the correct design preference, rather than reflecting generic principle coverage.

\subsection{Population-Level Alignment}
\label{sec:exp_calibration}

\begin{figure}[tbp]
\centering
\includegraphics[width=1.0\linewidth]{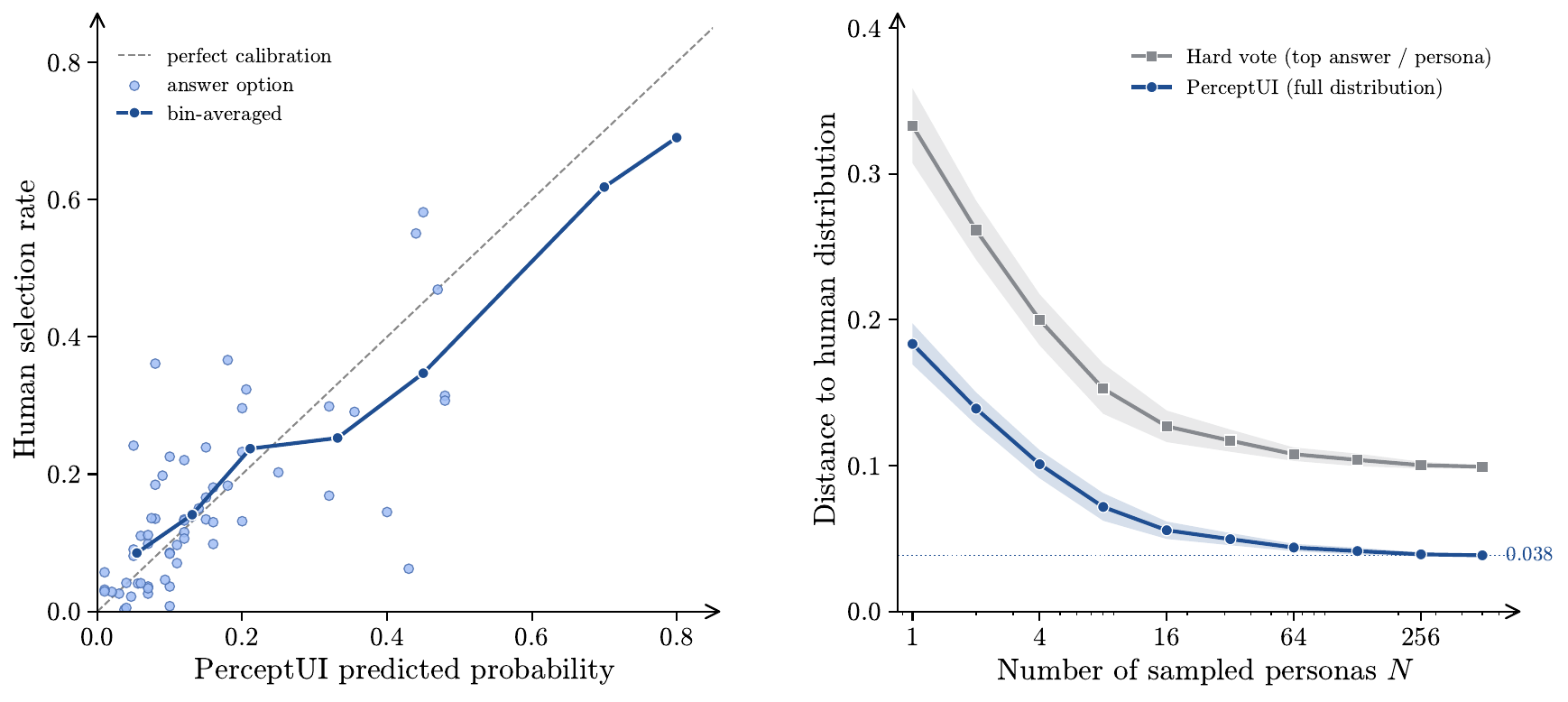}
\caption{Population-level calibration on UXcar. We compare predicted answer distributions with empirical human frequencies using JS divergence, a symmetric measure of distributional mismatch. Shaded regions denote one standard deviation over $20$ random subsamples.}
\label{fig:calibration}
\end{figure}

Beyond predicting individual answers, \textsc{PerceptUI} aims to estimate how a population would respond to a UI/UX question by aggregating persona-conditioned predictions. Figure~\ref{fig:calibration} evaluates this distributional calibration on UXcar, by aggregating answers over $N$ agents. The predicted answer probabilities are close to the observed human frequencies across answer options, although the model is slightly overconfident when assigning very high probabilities. The right figure also shows the effect of the persona budget $N$. We compare two aggregation methods. In hard voting, each persona contributes only its most likely answer. In soft aggregation, each persona contributes its full predicted answer distribution. Soft aggregation produces lower JS divergence because it preserves uncertainty from each persona instead of collapsing every prediction to a single choice. Increasing $N$ improves calibration, with gains beginning to saturate around $N{\approx}64$.

\subsection{Sensitivity to the Teacher Model}
\label{sec:exp_teacher_sensitivity}

\begin{figure}[tbp]
\centering
\includegraphics[width=1.0\linewidth]{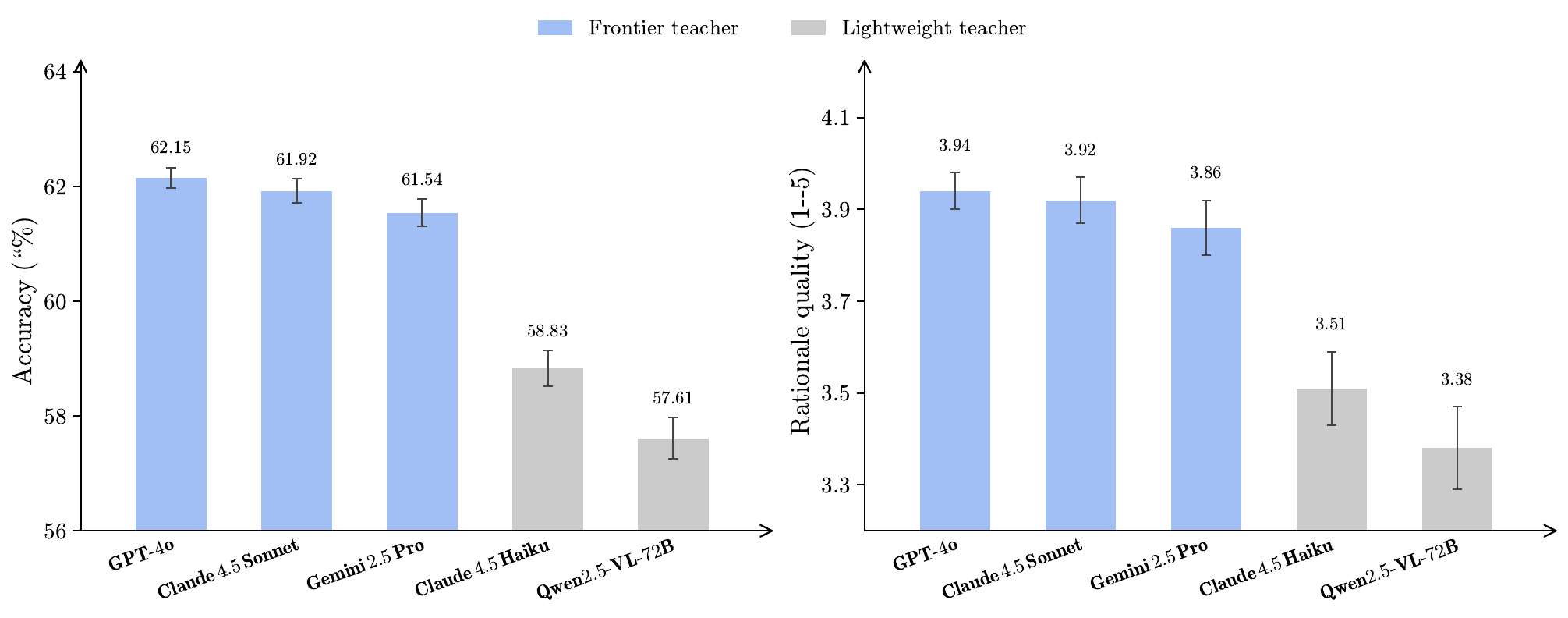}
\caption{Sensitivity to the teacher model used for contrastive reflection fine-tuning. The same student is fine-tuned with rationales from different teachers, and evaluated on answer accuracy and rationale quality.}
\label{fig:teacher_sensitivity}
\end{figure}
Contrastive reflection fine-tuning depends on rationales generated by a teacher model, so we test whether the gains are tied to the default teacher. We replace GPT-5.5 with four alternative teachers and keep the student model, training data, and prompt-evolution procedure fixed. Figure~\ref{fig:teacher_sensitivity} shows that frontier teachers yield similar downstream accuracy and rationale quality, with differences close to seed-level variation. This suggests that \textsc{PerceptUI} is relatively robust to the teacher choice. Performance drops when using smaller or weaker teachers, including Claude 4.5 Haiku and Qwen2.5-VL-72B, indicating that the quality of the generated contrastive rationales still matters. Overall, the results support the design of CRFT while showing that teacher choice affects the quality of the supervision signal.

\subsection{Generalization to Unseen Questions and Personas}
\label{sec:exp_generalization}

\begin{table}[tbp]
\centering
\small
\resizebox{0.95\linewidth}{!}{
\begin{tabular}{lcccc}
\toprule
\textbf{Method} & \textbf{Seen Q/P} & \textbf{Unseen Q} & \textbf{Unseen P} & \textbf{Unseen Q/P} \\
\midrule
No persona & 52.18 & 49.36 & 51.74 & 48.92 \\
Generic persona & 54.06 & 50.81 & 53.28 & 50.14 \\
Demographic persona & 56.34 & 52.27 & 54.86 & 51.63 \\
History-inferred persona & 58.61 & 53.80 & 56.22 & 52.40 \\
\midrule
Answer-only SFT & 56.72 & 52.61 & 55.18 & 51.94 \\
Positive-rationale SFT & 58.04 & 54.23 & 56.72 & 53.36 \\
\textsc{PerceptUI} w/o CR & 31.37 & 26.15 & 25.41 & 20.22 \\
\textsc{PerceptUI} w/o RPE & 60.37 & 56.15 & 58.41 & 55.22 \\
\rowcolor{blue!8}
\textsc{PerceptUI} & 62.15 & 58.34 & 60.26 & 57.08 \\
\bottomrule
\end{tabular}}
\caption{Generalization to unseen personas and questions on UXcar. Values report answer accuracy}
\label{tab:generalization}
\end{table}
This study assesses whether \textsc{PerceptUI} learns reusable user--UI judgment patterns rather than memorizing survey-specific questions or participant identities. We test three generalization settings: unseen questions, unseen participants, and the combination of both, allowing us to evaluate generalization to new question formulations and unseen user personas. As illustrated in Table~\ref{tab:generalization}, as expected, performance drops for all methods, with the largest degradation when both questions and participants are unseen. \textsc{PerceptUI} remains the strongest method across all settings, indicating that its gains are not limited to memorizing familiar users or question forms. The improvement is especially noticeable for unseen questions, where the model must interpret new wording while grounding its prediction in the UI and persona. On the other hand, the SFT baselines degrade more sharply, as imitating answers teaches the model to reproduce recorded answers without understanding why a choice fits a given user and UI.

\subsection{Visual Evidence Localization}
\label{sec:exp_uicrit_localization}
 \begin{figure}[tbp]
    \centering
    \includegraphics[width=1.0\linewidth]{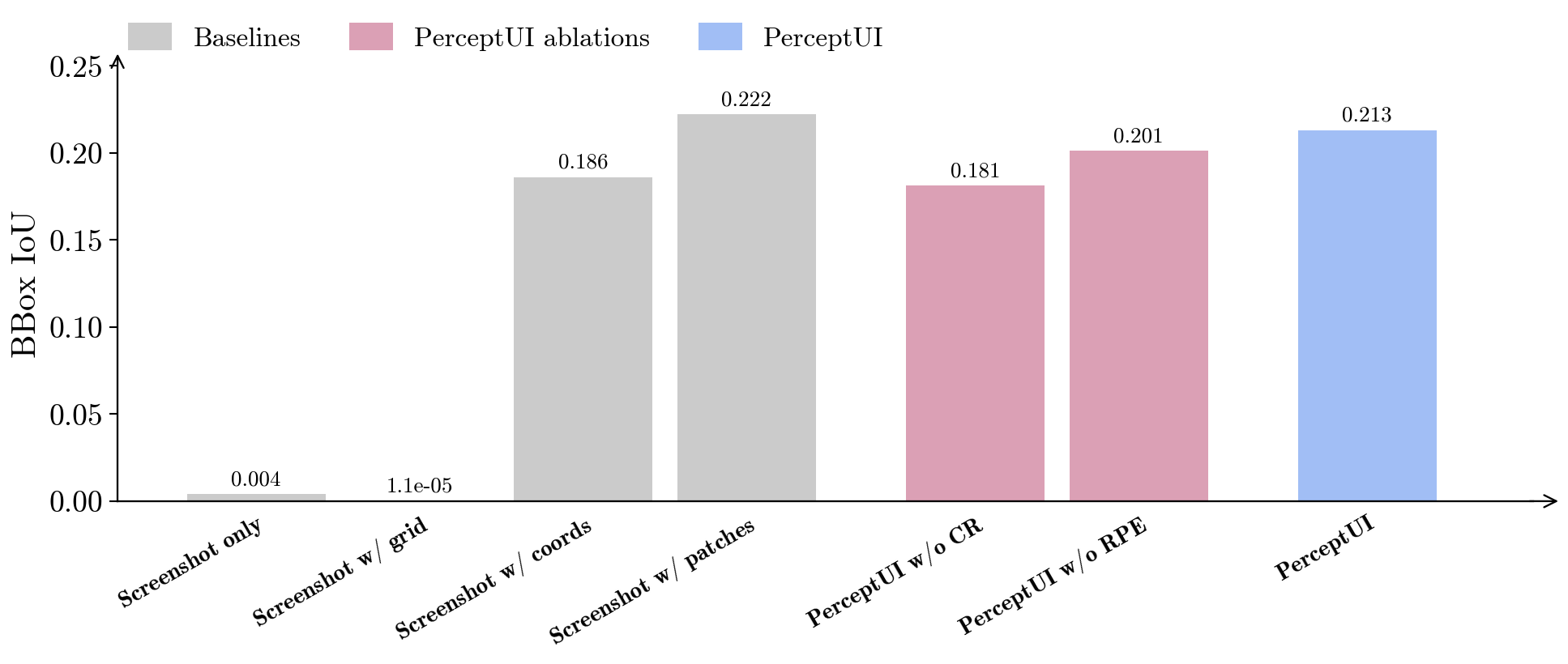}
    \caption{Visual evidence localization on UICrit (BBox IoU vs.\ expert annotations).}
    \label{fig:uicrit_localization}
  \end{figure}

Grounded UI critique requires not only identifying a design issue, but also locating the visual evidence that supports it. We evaluate this ability on UICrit~\citep{duan2024uicrit}, where the model outputs a bounding box for the UI region associated with its critique or predicted answer. Figure~\ref{fig:uicrit_localization} indicates that localization remains challenging for all methods. The strongest UICrit baseline relies on image patches and achieves the highest IoU among the original baselines. \textsc{PerceptUI} improves over coordinate prompting, but remains slightly below the patch-based baseline, which is expected because our model is optimized for answer prediction and explanation rather than dense visual grounding. The drop without contrastive reflection suggests that contrastive rationales also help the model attend to UI regions that distinguish the selected answer from alternatives.

\subsection{Evolution of Prompts}
\label{sec:exp_rpe_dynamics}

\begin{figure}[tbp]
\centering
\includegraphics[width=1.0\linewidth]{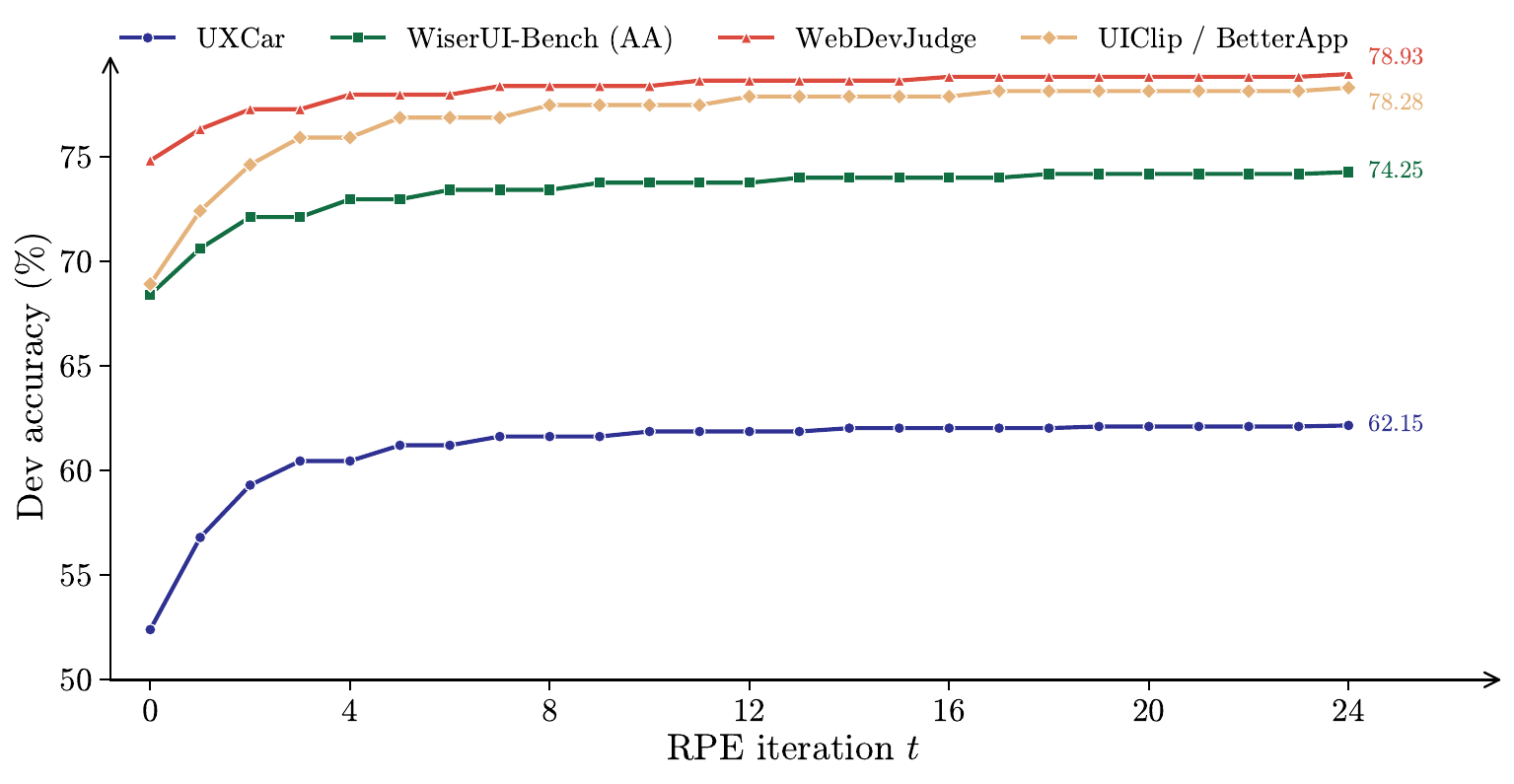}
\caption{Accuracy during reflective prompt evolution across four UI-evaluation benchmarks.}
\label{fig:rpe_evolution}
\end{figure}

Reflective prompt evolution is intended to adapt the inference prompt to recurring errors without changing the model parameters. To examine this process, each run starts from the same initial task-description prompt, and performance is tracked across optimization rounds. Figure~\ref{fig:rpe_evolution} indicates that most gains occur in the early iterations, followed by more gradual improvements. This suggests that early rewrites address broad prompt issues, such as ambiguous scale interpretation, weak use of persona information, or insufficient comparison between answer options. Later revisions mainly correct narrower errors and yield smaller marginal gains. The amount of improvement varies across benchmarks. The participant survey and UIClip/BetterApp benefit most, WiserUI-Bench improves more moderately, and WebDevJudge changes least. This pattern is consistent with the role of prompt evolution: it is most effective when errors arise from task framing or interpretation, and less effective when the relevant evidence is not fully available in the input.

\subsection{Ablation Study}
\label{sec:exp_ablation}

\begin{table}[tbp]
\centering
\small
\resizebox{0.98\linewidth}{!}{
\begin{tabular}{lcccc}
\toprule
\textbf{Method} & \textbf{Acc.}$\uparrow$ & \textbf{Macro-F1}$\uparrow$ & \textbf{JS Div.}$\downarrow$ & \textbf{Rationale}$\uparrow$ \\
\midrule
Majority class & 39.62 & 28.41 & 0.161 & -- \\
SFT & 48.93 & 39.30 & 0.112 & 2.74 \\
\midrule
No persona & 52.18 & 45.37 & 0.074 & 2.84 \\
Shuffled persona & 52.06 & 45.11 & 0.076 & 2.81 \\
Generic persona & 54.06 & 47.12 & 0.066 & 3.03 \\
\midrule
\textsc{PerceptUI} w/o CR & 31.37 & 24.07 & 0.231 & 1.89 \\
\textsc{PerceptUI} w/o RPE
& \cellcolor{wiseryellow}\underline{60.37}
& \cellcolor{wiseryellow}\underline{53.12}
& \cellcolor{wiseryellow}\underline{0.048}
& \cellcolor{wiseryellow}\underline{3.71} \\
\rowcolor{blue!8}
\textsc{PerceptUI}
& \cellcolor{wiserblue}\textbf{62.15}
& \cellcolor{wiserblue}\textbf{55.04}
& \cellcolor{wiserblue}\textbf{0.039}
& \cellcolor{wiserblue}\textbf{3.94} \\
\bottomrule
\end{tabular}}
\caption{Ablation study on UXcar. Accuracy and Macro-F1 evaluate answer prediction, JS divergence measures distributional calibration against the human answer distribution (lower is better), and Rationale is the average expert-rated explanation quality on a 1--5 scale.}
\label{tab:ablation}
\end{table}
We ablate the main components of \textsc{PerceptUI} on UXcar. We first examine the role of persona information by removing the persona, replacing it with a shuffled participant profile, or using a shared generic description. We then isolate the effect of the learning and inference stages by comparing answer-only fine-tuning (SFT), contrastive reflection fine-tuning, and reflective prompt evolution. Table~\ref{tab:ablation} shows that persona variants provide only limited gains when the profile is generic or mismatched, suggesting that participant-level prediction requires user context that is both specific and correctly grounded. Answer-only fine-tuning also remains limited, as it learns from target labels without modeling why an answer is preferred over alternatives. In contrast, contrastive reflection fine-tuning yields the largest improvement, indicating that rationales grounded in the UI, persona, and answer options provide useful supervision for learning user-specific judgment cues. 

\section{Examples}

\begin{tcolorbox}[colframe=customblues3, colback=white, title={Worked Example: Auxiliary Groundings $(u_i, g_i)$}, breakable ] \textbf{UI grounding $u_i$.} The dashboard shows the vehicle traveling at 67\,km/h with a posted speed limit of 100 and a set speed of 80. A lane-assist visualization displays the car centered in its lane with blue guidance ahead, while the navigation panel on the right highlights an upcoming curved right lane / exit assistance prompt in 700\,m. Green driver-assistance and HOLD indicators are visible, and the road scene ahead appears clear with vehicles shown farther in front.\\[4pt] \textbf{Persona grounding $g_i$.} The participant is a regular U.S.\ driver with consistent car access and some monthly experience using adaptive cruise control, so he is likely familiar with driver-assistance prompts and related terminology. His hands-on domain experience suggests that he may recognize specialized vehicle icons, though his neutral / noncommittal tendency may make him cautious if the screen message is ambiguous. \end{tcolorbox}

\begin{tcolorbox}[
  colframe=customblues3,
  colback=white,
  title={Contrastive Rationale $c_i^\star$},
  breakable
]
\textbf{UI evidence.} The screen shows active driver-assistance / lane guidance, with the car centered in a blue lane path and a steering-wheel assist icon at the bottom. The right-side panel also highlights an upcoming assisted maneuver / curve in 700\,m, while there is no visible instruction indicating that hands-free driving is allowed. These elements support keeping or putting hands on the steering wheel while the assistance system is active.\\[4pt]
\textbf{Persona relevance.} As a regular driver with monthly ACC experience and hands-on familiarity with driving-assistance terminology, the participant would likely interpret the assist graphics as requiring driver supervision rather than full automation. His neutral, cautious tendency also makes ``put your hands on the steering wheel'' the natural choice when the screen is not explicitly telling him to go hands-free.\\[4pt]
\textbf{Option contrast.}
\begin{itemize}
  \item \textbf{(1)} Taking hands off the steering wheel would feel less appropriate because the display shows assistance, not a clear hands-free authorization.
  \item \textbf{(3)} ``Not sure / don't know'' would be less appropriate because his driving and ACC experience give him enough context to treat the steering-wheel assist display as a prompt to stay engaged.
\end{itemize}
\end{tcolorbox}

\end{document}